\documentclass[journal=tches]{iacrtrans}

\usepackage{algorithm}
\usepackage{algorithmic}
\usepackage{booktabs} 
\usepackage{caption}
\usepackage{cite}
\usepackage{csquotes}
\usepackage{comment}
\usepackage[para]{footmisc}
\usepackage{hyperref}
\hypersetup{
    citecolor=darkblue
}
\usepackage{pifont}
\usepackage{makecell}
\usepackage{mathtools}
\usepackage{multirow}
\usepackage[autolanguage]{numprint}
\usepackage{orcidlink}
\usepackage{rotating}
\usepackage{subfigure}
\usepackage{threeparttable}
\usepackage{todonotes}
\usepackage{url}            
\usepackage{xcolor}         
\usepackage{xspace}

\definecolor{codegreen}{rgb}{0,0.6,0}
\definecolor{codegray}{rgb}{0.5,0.5,0.5}
\definecolor{codepurple}{rgb}{0.58,0,0.82}
\definecolor{backcolour}{rgb}{0.95,0.95,0.92}
\usepackage{listings}

\lstdefinestyle{python}{ 
  basicstyle=\ttfamily\scriptsize,
  backgroundcolor=\color{backcolour},
  keywordstyle=\color{magenta},
  stringstyle=\color{red},
  commentstyle=\color{green},
  frame=single,
  breaklines=true,
  breakatwhitespace=true,
  showstringspaces=false,
  captionpos=b,
  numbers=left,
  numberstyle=\tiny\color{gray},
  stepnumber=1,
  numbersep=5pt,
  linewidth=\textwidth
}

\newcommand\algorithmicinput{\textbf{Input:}}
\newcommand\INPUT{\item[\algorithmicinput]}

\newcommand\algorithmicoutput{\textbf{Output:}}
\newcommand\OUTPUT{\item[\algorithmicoutput]}

\newcommand\algorithmicactors{\textbf{Actors:}}
\newcommand\ACTORS{\item[\algorithmicactors]}
\newcommand{\FLname}{WW\mbox{-}FL\xspace}
\newcommand{\FLnameTitle}{WW-FL}

\definecolor{darkred}{rgb}{.6,0,0}
\definecolor{darkgreen}{rgb}{0,.4,0}
\definecolor{darkblue}{rgb}{0,0,.6}

\definecolor{absolutezero}{rgb}{0.0,0.28,0.73}
\definecolor{acidgreen}{rgb}{0.69,0.75,0.1}
\definecolor{amazon}{rgb}{0.23,0.48,0.34}
\definecolor{ambersaeece}{rgb}{1.0,0.49,0.0}
\definecolor{antiquebronze}{rgb}{0.4,0.36,0.12}
\definecolor{aoenglish}{rgb}{0.0,0.5,0.0}
\definecolor{brightmaroon}{rgb}{0.76,0.13,0.28}
\definecolor{cadmiumred}{rgb}{0.89,0.0,0.13}
\definecolor{britishracinggreen}{rgb}{0.0,0.26,0.15}
\definecolor{lightgray}{gray}{0.9}

\newcommand{\secref}[1]{\S\ref{#1}}
\newcommand{\figref}[1]{Fig.~\ref{#1}}
\newcommand{\tabref}[1]{Tab.~\ref{#1}}
\newcommand{\algref}[1]{Alg.~\ref{#1}}
\newcommand{\alglineref}[1]{\ensuremath{{\sf L}\ref{#1}}}

\newcommand{\makeparafit}{\looseness=-1}

\newcommand{\cmark}{\textcolor{darkgreen}{\ding{51}}}%
\newcommand{\xmark}{\textcolor{darkred}{\ding{55}}}%

\newcommand{\Adv}{\ensuremath{\mathcal{A}}} 
\newcommand{\server}{\ensuremath{\mathcal{S}}}
\newcommand{\client}{\ensuremath{\mathcal{C}}}

\newcommand{\partyset}{\ensuremath{\mathcal{P}}}

\newcommand{\globalserverset}{\ensuremath{\mathcal{G}}}
\newcommand{\globalserver}[1]{\ensuremath{\mathsf{GS}_{#1}}}
\newcommand{\clusterserverset}[1]{\ensuremath{\mathcal{M}_{#1}}}
\newcommand{\clusterserver}[2]{\ensuremath{\mathsf{CS}_{#1}^{#2}}} 
\newcommand{\clientFLset}[1]{\ensuremath{\mathcal{C}_{#1}}}
\newcommand{\clientFL}[2]{\ensuremath{\mathsf{C}_{#1}^{#2}}} 
\newcommand{\clientFLsetTotal}[1]{\ensuremath{\mathcal{C}_{#1}^{\sf U}}}
\newcommand{\sizeFL}[1]{\ensuremath{\mathsf{N}_{#1}}}
\newcommand{\globalserversize}{\ensuremath{\sizeFL{\globalserverset}}}
\newcommand{\clusterserversize}[1]{\sizeFL{\clusterserverset{#1}}}
\newcommand{\clusterclientsize}[1]{\sizeFL{\clientFLset{#1}}}
\newcommand{\numclusters}{\sizeFL{\scriptstyle \mathcal{M}}}
\newcommand{\SA}{\textnormal{\textsc{SA}}}
\newcommand{\singleserverSA}{\textnormal{\textsc{Single~$\server$}}}
\newcommand{\multiserverSA}{\textnormal{\textsc{Multi~$\server$}}}

\newcommand{\AdvStrategy}[1]{\ensuremath{\mathsf{St}_{#1}}} 
\newcommand{\entityset}{\ensuremath{\gamma}}
\newcommand{\adventityset}{\ensuremath{\delta}}
\newcommand{\MPCShare}{\textnormal{\textsc{Share}}}
\newcommand{\MPCAgg}{\textnormal{\textsc{Agg}}}
\newcommand{\MPCTrain}{\textnormal{\textsc{Train}}}
\newcommand{\MPCInf}{\textnormal{\textsc{Predict}}}
\newcommand{\MPCReshare}{\textnormal{\textsc{Reshare}}}
\newcommand{\MPCReveal}{\textnormal{\textsc{Reveal}}}
\newcommand{\MPCSample}{\textnormal{\textsc{Sample}}}

\newcommand{\MPCTrimmedMeanList}{\textnormal{\textsc{TM-List}}}

\newcommand{\MPCTopKHitter}{\textnormal{\textsc{TopK-Hitter}}}
\newcommand{\FLTrain}{\textnormal{\textsc{FL}\mbox{-}\textsc{Train}}}

\newcommand{\myMPCSet}[1]{\ensuremath{\mathcal{#1}}}
\newcommand{\MPCIndexList}{\ensuremath{\mathcal{I}}}

\newcommand{\MPCData}[2]{\ensuremath{{D}_{#1}^{#2}}}
\newcommand{\MPCModel}[2]{\ensuremath{{W}_{#1}^{#2}}}
\newcommand{\MPCQuery}[2]{\ensuremath{{Q}_{#1}^{#2}}}

\newcommand{\MPCSource}{\ensuremath{{U}}}
\newcommand{\MPCTarget}{\ensuremath{{V}}}

\newcommand{\MPCSharing}[2]{\ensuremath{{\langle#1\rangle}_{#2}}}

\newcommand{\func}[1]{\ensuremath{\mathcal{F}_{#1}}}
\newcommand{\funcMPC}{\func{\textsf{MPC}}}
\newcommand{\funcFLname}{\func{\textsf{\FLname}}}
\newcommand{\funcShare}{\func{\MPCShare}}
\newcommand{\funcAgg}{\func{\MPCAgg}}
\newcommand{\funcTrain}{\func{\MPCTrain}}
\newcommand{\funcReshare}{\func{\MPCReshare}}
\newcommand{\funcSample}{\func{\MPCSample}}

\newcommand{\TRParam}{\ensuremath{{\alpha}}}
\newcommand{\TRSampleSize}{\ensuremath{{\beta}}}
\newcommand{\FLmodelSet}[1]{\ensuremath{\mathcal{W}^{#1}}}
\newcommand{\FLmodelSize}{\ensuremath{{\gamma}}}
\newcommand{\OutlierSet}{\ensuremath{\mathcal{Z}}}
\newcommand{\FLmodel}[2]{\ensuremath{W_{#2}^{#1}}} 
\newcommand{\WWFLmodel}[1]{\ensuremath{W_{#1}}} 
\newcommand{\WWFLData}[2]{\ensuremath{{D}_{#1}^{#2}}}

\newcommand{\mv}[1]{\ensuremath{{\sf m}_{#1}}}
\newcommand{\lv}[2]{\ensuremath{{\lambda}_{#1}^{#2}}}

\newcommand{\Falcon}{\textsc{Falcon}\xspace}
\newcommand{\Spindle}{\textsc{Spindle}\xspace}
\newcommand{\Poseidon}{\textsc{Poseidon}\xspace}
\newcommand{\Hercules}{\textsc{Hercules}\xspace}

\newenvironment{proofsketch}{\paragraph{Proof Sketch}}{\hfill$\square$}

\newcommand{\myparagraph}[1]{\vspace{-3mm}\paragraph{#1}}

\newenvironment{tableitem}{
	\begin{list}{{$-$}}{
			\setlength\partopsep{0pt}
			\setlength\parskip{0pt}
			\setlength\parsep{2pt}
			\setlength\topsep{0pt}
			\setlength\itemsep{0pt}
			\setlength{\itemindent}{5pt}
			\setlength{\leftmargin}{0pt}
		}
	}{
	\end{list}
}

\newenvironment{myitemize}{
	\begin{list}{{$\bullet$}}{
			\setlength\partopsep{0pt}
			\setlength\parskip{0pt}
			\setlength\parsep{2pt}
			\setlength\topsep{4pt}
			\setlength\itemsep{3pt}
			\setlength{\itemindent}{15pt}
			\setlength{\leftmargin}{2pt}
		}
	}{
	\end{list}
}

\newenvironment{compactitemize}{
	\begin{list}{{$\bullet$}}{
			\setlength\partopsep{0pt}
			\setlength\parskip{0pt}
			\setlength\parsep{2pt}
			\setlength\topsep{2pt}
			\setlength\itemsep{2pt}
			\setlength{\itemindent}{15pt}
			\setlength{\leftmargin}{2pt}
		}
	}{
	\end{list}
}

\newenvironment{myitem}{
	\begin{list}{{$\bullet$}}{
			\setlength\partopsep{0pt}
			\setlength\parskip{0pt}
			\setlength\parsep{2pt}
			\setlength\topsep{4pt}
			\setlength\itemsep{3pt}
			\setlength{\itemindent}{20pt}
			\setlength{\leftmargin}{2pt}
		}
	}{
	\end{list}
}

\author{
    Felix~Marx\inst{1}\orcidlink{0000-0003-3899-3258} \and 
    Thomas~Schneider\inst{1}\orcidlink{0000-0001-8090-1316} \and
    Ajith~Suresh\inst{2}\orcidlink{0000-0002-5164-7758} \and 
    Tobias~Wehrle\inst{1}\orcidlink{0000-0002-1316-1001} \and 
    Christian~Weinert\inst{3}\orcidlink{0000-0003-4906-6871} \and
    Hossein~Yalame\inst{1}\orcidlink{0000-0001-6438-534X} 
}
\institute{
  Technical University of Darmstadt, Germany\\
  \email[
      felix_marx@t-online.de,
      schneider@encrypto.cs.tu-darmstadt.de,
      tobi-wehr@web.de,
      yalame@encrypto.cs.tu-darmstadt.de
  ]{{lastname}@encrypto.cs.tu-darmstadt.de}
  \and
  Technology Innovation Institute, Abu Dhabi, UAE\\
  \email[
      ajith.suresh@tii.ae
  ]{{firstname.lastname}@tii.ae}
  \and
  Royal Holloway, University of London\\
  \email[
      christian.weinert@rhul.ac.uk
  ]{{firstname.lastname}@rhul.ac.uk}
}
\authorrunning{F.Marx et al.}

\title{\FLnameTitle: Secure and Private Large-Scale Federated Learning (Full Version)\thanks{Please cite the journal version of this work, to be published in IACR Transactions on Cryptographic Hardware and Embedded Systems (CHES’26)~\cite{CHES:MSSWWY26}}}

\begin{document}

\maketitle


\begin{abstract}
Federated learning~(FL) is an efficient approach for large-scale distributed machine learning that promises data privacy by keeping training data on client devices. However, recent research has uncovered vulnerabilities in~FL, impacting both security and privacy through poisoning attacks and the potential disclosure of sensitive information in individual model updates as well as the aggregated global model. This paper explores the inadequacies of existing~FL protection measures when applied independently, and the challenges of creating effective compositions. 

Addressing these issues, we propose~\FLname{}, an innovative framework that combines secure multi-party computation~(MPC) with hierarchical~FL to guarantee data and global model privacy. One notable feature of~\FLname{} is its capability to prevent malicious clients from directly poisoning model parameters, confining them to less destructive data poisoning attacks. We furthermore provide a PyTorch-based~FL implementation integrated with Meta's CrypTen MPC framework to systematically measure the performance and robustness of~\FLname{}. Our extensive evaluation demonstrates that~\FLname{} is a promising solution for secure and private large-scale federated learning.
\end{abstract}
\section{Introduction}
\label{sec:intro}
Federated learning~(FL)~\cite{AISTATS:McMahanMRHA17} is a revolutionary approach for large-scale distributed machine learning that emerged in the past decade and has had a profound impact on both academia~\cite{csur:YinZH21} and industry~\cite{NPJDM:RiekeH0MRABGLMO20,ARXIV:ibmfl2020ibm,ARXIV:FEDML}. A World Wide Web Consortium~(W3C) community group currently works towards establishing~FL-related~Web standards~\cite{WEB:W3C}.
FL shares some objectives with privacy-preserving machine learning~(PPML)~\cite{ICML:Gilad-BachrachD16,SP:MohasselZ17,NgC23,CHES:SanderBBE25}, but differs in its approach to training a model: In~PPML, training on the~\emph{entire set of available data} is either outsourced to a remote party or done collaboratively while ensuring data privacy. In contrast, FL involves multiple rounds of training, where each round consists of selected clients~\emph{locally training a model on their private data} and a central party performing aggregation of these models to update a global model.

Initially, FL appeared to indeed offer privacy since the training data remains on the client devices, and only model updates~(so-called gradients) are transmitted. However, subsequent research has demonstrated that these gradients still reveal a significant amount of information, enabling the central aggregator to infer sensitive details about the clients' private training data~\cite{NIPS:GeipingBD020,SP:ZSEEAB24,ICLR:WBSCZXS24,USENIX:DHWSGRC25}. To address this issue, \emph{secure aggregation}~(SA) schemes~\cite{CCS:BIKMMP17,CCS:BellBGL020,PETS:MOJC23} were proposed, where only the aggregated result is revealed to the central party, effectively concealing the individual gradients.
Despite this, FL still poses several complex challenges that demand careful consideration~\cite{CSUR:ZBACCDLLNR25}.

The first challenge relates to serious privacy vulnerabilities that have been identified when using secure aggregation with a \emph{single} aggregator~\cite{WEB:BDSSSP22,ICLR:FowlGCGG22,ICML:WenGFGG22,AAAI:SAGJA23,EUROSP:BDSSSP23,SP:ZSEEAB24}: Recent studies have demonstrated the effectiveness of privacy attacks even when utilizing secure aggregation techniques across many clients, showing that such schemes are insufficient~\cite{CCS:PFA22,ARXIV:BDSSSP23}.
These vulnerabilities largely stem from the imbalance of power held by the central aggregator, who can introduce inconsistencies in the global model, manipulate client selection, and exert disproportionate influence over the training process.

These growing concerns over privacy with a single central aggregator hinder the use of standard~FL, and even~FL with~SA, for global  training tasks as sharing sensitive data across jurisdictions is often problematic legally.
For example, the~Court of Justice of the European Union~(CJEU) with its famous~\enquote{Schrems~II} judgment has ruled the~\enquote{Privacy Shield} agreement between the~EU and the~US for exchanging personal data invalid~\cite{tracol2020schrems}. Until agreeing on the most recent~\enquote{Data Privacy Framework}~\cite{DPF}, numerous attempts to set up data sharing have failed, including the~\enquote{Safe Harbour} principles~\cite{weiss2016us}.
Recent works therefore suggest a distributed aggregator setup using secure multi-party computation~(MPC) to securely distribute aggregation tasks among multiple servers, ensuring privacy despite potential collusion among a subset of them~\cite{DLS:FereidooniMMMMN21,SP:RatheeSWP23,DLS:GMSSWY23,SATML:BMPSTVWYY24,TCHES:TXLZ25}.

The second~(and closely related) challenge is the current lack of~\emph{global model privacy}.
Initially, FL focused on enhancing user participation and improving model accuracy rather than safeguarding the model itself.
However, with the widespread adoption of~FL in various industries, such as the medical sector~\cite{NPJDM:RiekeH0MRABGLMO20}, the need for privacy protection becomes apparent.
Unfortunately, unrestricted access to the aggregated model allows extraction of traces of original training data, necessitating~\emph{global model privacy}~\cite{CCS:PFA22,ARXIV:BDSSSP23,SP:ZSEEAB24,INFOCOM:WCHPDCR24}.
So far, only a small number of works have focused on preserving the privacy of the global model in~FL~\cite{CCSW:MandalG19,NDSS:SavPTFBSH21}.
Most of these works use homomorphic encryption~(HE) schemes~\cite{CSURV:carAUC18}, which primarily address semi-honest corruptions.
Moreover, several works do not consider the collusion of clients with the aggregator and thereby provide a weaker notion of security.
Furthermore, organizations coordinating FL may have a vested interest in keeping the global model private---even from the contributing clients---as revealing the model could pose business risks~\cite{EMNLP:DXDSL23}.

The third challenge arises when corrupted clients intentionally manipulate the locally trained model to reduce the system's accuracy or introduce backdoors to extract sensitive information in future rounds~\cite{NIPS:WangSRVASLP20,ICLR:XieHCL20,ICML:ZPSYMMRG22}.
This method of~\emph{model poisoning} is more potent than~\emph{data poisoning}~\cite{SP:SHKR22,ARXIV:XFG24}, which involves manipulating the dataset.
Although numerous defenses have been proposed against model poisoning~\cite{NDSS:ShejwalkarH21,NDSS:CaoF0G21,KDD:ZCJG22}, the most effective remain impractical for large-scale deployment due to their high computation and communication costs~\cite{USENIX:NguyenRCYMFMMMZ22,SP:RatheeSWP23,TCHES:TXLZ25}.
Additionally, recent research has introduced even more potent model poisoning strategies that can circumvent state-of-the-art defenses~\cite{NDSS:ShejwalkarH21,CSUR:GGP24,ARXIV:XFG24}. 
For instance, PoisonedFL~\cite{ARXIV:XFG24} breaks eight state-of-the-art defenses without any knowledge of the honest clients' models, highlighting the urgent need for new defense mechanisms.
To address corrupted clients, we limit their attack capabilities to data poisoning rather than the more severe model poisoning. This allows us to implement more moderate defenses compared to existing schemes~(cf.~\secref{sec:attacks_defenses} for details).

Besides these three challenges, a large-scale, worldwide FL deployment needs to account for client heterogeneity in terms of bandwidth, computational power, jurisdiction, and trust assumptions.
As we will elaborate in~\secref{sec:motivation}, trust in government and private entities varies widely, influencing how data is shared and used.

Although various works have addressed the aforementioned issues individually or in combination, a comprehensive solution remains elusive. Therefore, we propose a unified framework called \FLname{}, which enables secure and private distributed machine learning at scale. Departing from traditional~FL, we introduce the concept of a \emph{trust zone}, allowing data to be shared securely within a trusted legal jurisdiction and leveraging privacy-enhancing technologies like~MPC.
This approach supports edge nodes with limited resources and encourages wider participation, enhancing the system's overall privacy and performance.

\subsection{A Regulatory and Trust Perspective} 
\label{sec:motivation}
Beyond the critical security, privacy, and technical deployment challenges that exist in~FL and have been discussed so far, it is equally important to consider the regulatory landscape and the varying degrees of trust among participating entities. Specifically,~\FLname{} aims to address the~\emph{heterogeneity in trust} across entities distributed geographically~\cite{Book:TrustOCED,WEB:TrustCountry}.
This variation is not only regional but also institutional, with differing levels of trust in, for example, law enforcement versus judicial bodies~\cite{Book:TrustOCED}.

When it comes to private companies, some countries have implemented strict laws to ensure the ethical behavior of entities operating within their borders.
Non-compliance can result in substantial fines and even imprisonment~\cite{WEB:GDPRFines,WEB:BigBreachFines}.
Recent examples include~Meta receiving a~1.2 billion~Euro fine from regulators in~Ireland~\cite{WEB:GDPRFines} and~Didi Global facing a~1.19 billion~Euro penalty from the~Cyberspace Administration of China~\cite{WEB:BigBreachFines}. 
Several companies including well-known~TalkTalk and~Target either lost their customers' data in major data breaches or were accused of not keeping customer information safe. This led to a lack of trust among their clients and a significant drop in revenue~\cite{WEB:DataPrivacy,WEB:CostOfBreach}.

Given this disparity in trust, consider a consortium comprising three companies interested in training a machine learning~(ML) model for medical diagnostics.
Numerous studies have demonstrated the effectiveness of~ML models in diagnosing various diseases, such as lung cancer, tumor detection, and liver segmentation~\cite{IOTJ:RHJHBRV24}. However, these studies emphasize the need for diverse patient data.
Simultaneously, the sensitivity and privacy of the training data must be preserved.

To address this, the consortium may collaborate with various federal and local government agencies to obtain relevant training data and potentially compensating data providers. 
However, since the model is trained on sensitive medical data, the consortium members do not want the trained model to be disclosed to anyone.
Additionally, the consortium should have the ability to use the trained model for inference while preserving privacy.
This approach not only enhances the system's privacy by safeguarding against breaches that might affect individual companies but also safeguards the financial investments of these companies by preventing unauthorized use of their trained model by third parties.

At first glance, implementing a~PPML solution may appear simplistic.
In~PPML, data owners securely distribute their data among the three companies within the consortium, ensuring that no individual company can access the data in the clear.
The consortium employs~MPC techniques to privately train a model on the shared data.
However, this approach has several drawbacks.
One significant drawback is the substantial computational power required within the consortium to handle the training process, particularly when dealing with millions of data owners in a global deployment.
Furthermore, this approach necessitates a high level of trust from all data owners, as they must share their data among the consortium's member companies.
This trust requirement can be a significant hurdle to adoption. 
Additionally, there may be situations where data owners are willing to share their data but cannot do so due to government regulations preventing data from leaving their jurisdiction. 
This restriction not only limits data availability but also renders this approach unsuitable for our specific use case.

The next approach we can consider is using a traditional~FL with secure aggregation.
While this method has the potential to address the scalability problem, it comes with a significant drawback. In this approach, the model during training is distributed to the data owners. Although there are works like~\cite{PETS:FTPSSB21,NDSS:SavPTFBSH21,TDSC:XuHXZLHD23}, which protect the privacy of the global model using homomorphic encryption techniques, they may not be well-suited for our cross-device setting with a large number of data owners.
Additionally, these techniques necessitate interaction among the clients and require extra measures for handling dropouts.

In light of the limitations of current methods, we make an adjustment to the concept of~FL.
Specifically, we expand the notion of a~\emph{trust zone}, allowing data to be moved from the data owner to this larger, still-trusted area.
For instance, a user in our scenario can securely share their data among three entities: the police, the court, and a private company, all operating within the same legal jurisdiction as the data owner.
These selected entities will collaboratively train the model using privacy-enhancing technologies like~MPC and~HE. 
Furthermore, this approach has the added benefit of supporting edge nodes with limited computational resources, allowing them to contribute to the system.
Importantly, since data owners can be assured that their data remains within their designated trust zone, this will likely encourage more users to participate, leading to a larger pool of training data and, ultimately, a better-trained model while guaranteeing global privacy.

Finally, it is important to note that the concept of a trust-zone can differ depending on the specific region or context.
For example, in one country, it may be relatively straightforward to identify three entities that data owners trust not to collude with each other.
In contrast, there may be situations where it is challenging to find three or four entities that can provide an absolute guarantee of non-collusion.
However, in such cases, data owners may be willing to accept a weaker assurance that not all entities will collude.
We can accommodate this heterogeneity in trust by performing~PPML training independently for each of these scenarios, using the appropriate~MPC protocol, and then combining their results through secure aggregation using global servers.

\subsection{Our Contributions}
\begin{figure*}[htb!]
    \centering
    \includegraphics[width=\textwidth, trim={0 1cm 0 0.25cm},clip]{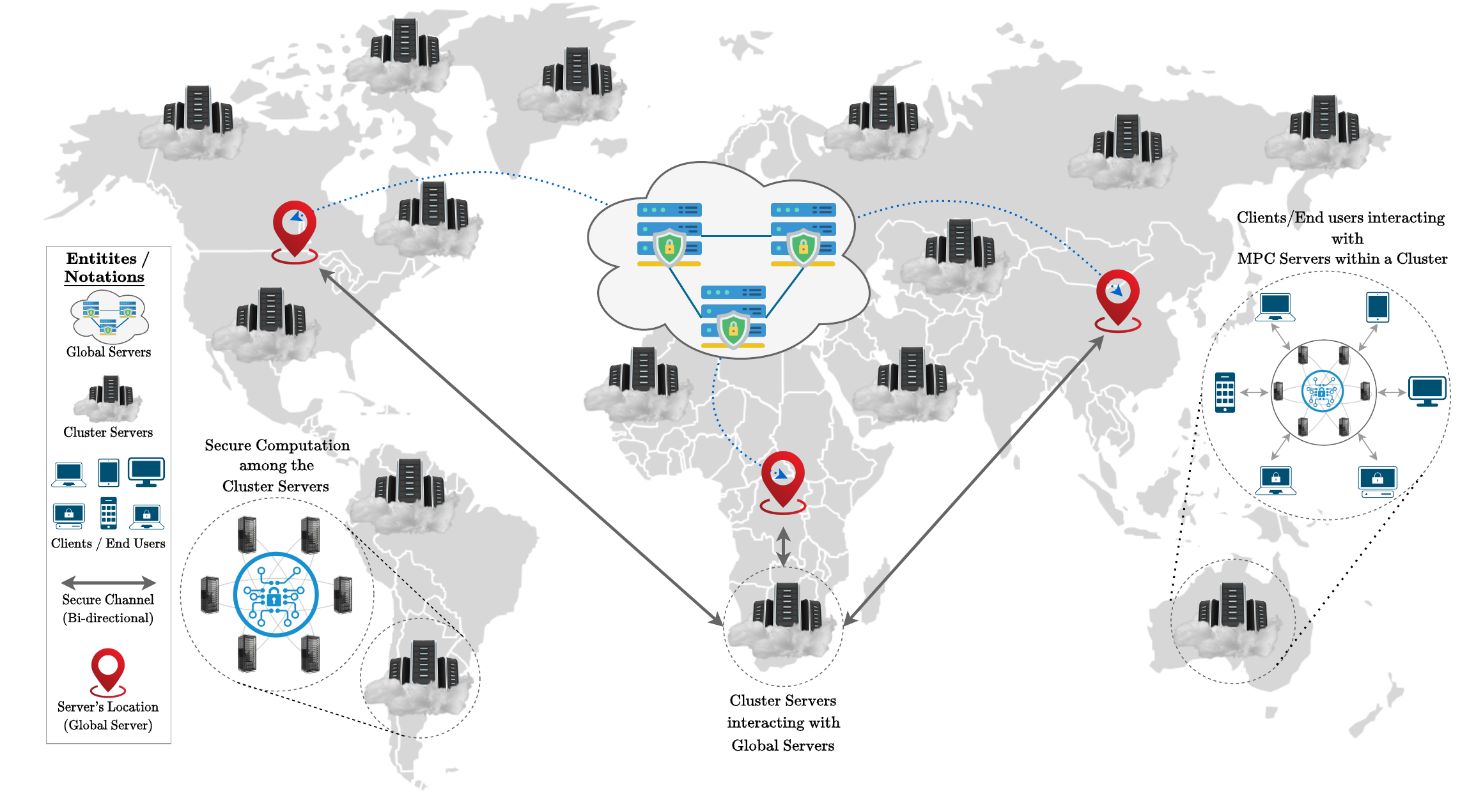}
    \vspace{-4mm}
    \captionsetup{font=small}
    \caption{Visual representation of deploying our secure world-wide~FL~(\FLnameTitle) framework. In this illustration, clients connect to local~MPC clusters, which in turn are connected to three global servers (this can be varied as per requirements) strategically positioned across continents. \figref{fig:architecture} provides a detailed illustration of \FLnameTitle, showcasing the roles of involved entities.}
    \label{fig:worldwide}
    \vspace{-2mm}
\end{figure*}

In this paper, we initiate a discussion on novel approaches towards secure and private large-scale~(cross-device) federated learning to address the challenges outlined above. Towards this goal, we propose a unified framework called~\enquote{worldwide-FL}~(\FLname). Our framework is based on a novel abstraction that also captures existing hierarchical~FL architectures in a~\emph{hybrid} manner as visualized in~Fig.~\ref{fig:worldwide} and detailed in~\secref{sec:framework-abstraction}.

Briefly, in \FLname, clients use secret-sharing to securely outsource training data to distributed~\emph{cluster servers} based on~MPC.
The clients then can leave and return sporadically to provide more training data -- this makes our framework robust against real-world issues like drop-outs, requires no interaction between clients, and resource-constrained~(mobile or edge) devices only have very little \mbox{workload}.
The models trained by the cluster servers with~MPC-based~PPML techniques are then aggregated across training clusters using one or multiple levels of distributed aggregators. For secure distributed aggregation, we again utilize~MPC.
After aggregation, models are not converted to plaintext but are returned to the training clusters for the next training iteration in secret-shared form.
When training is completed, known secure inference protocols can be used to allow private queries~\cite{NgC23,EPRINT:MWCB22} in a controlled~(potentially rate-limited) way.
Our architecture addresses issues with single aggregators by relying on a \emph{distributed aggregator} and notably achieves~\emph{global model privacy} due to a novel combination of~FL with~PPML techniques.

\pagebreak
 
We observe that a unique property of our framework is the strictly limited attack surface: similar to~PPML, malicious clients are restricted to weaker data-poisoning attacks as there is no possibility to access and manipulate the model itself.
As defending against data poisoning in~PPML is highly non-trivial, we show that state-of-the-art data-poisoning attacks in the suggested hierarchical configuration are less effective than in plain~FL. 
Furthermore, we consider different robust aggregation schemes to further mitigate the effect of such attacks. For this, we additionally propose new heuristics that improve the efficiency in~MPC. Specifically, we propose a lightweight variant of the~\enquote{Trimmed Mean}~\cite{ICML:YinCRB18} defense for our setting with only data-poisoning attacks.

\makeparafit
Finally, we systematically evaluate possible instantiations of \FLname in terms of performance~(both in terms of achievable accuracy as well as overhead due to~MPC) and resilience against poisoning attacks. For this, we implement a prototype based on~PyTorch and~Meta's CrypTen framework~\cite{NeurIPS:KnottVHSIM21} for instantiating~MPC. 
We then evaluate the performance when training neural networks for standard image classification tasks in realistic network settings and using~GPU-accelerated~AWS EC2 instances.
This way, we show that the computation and communication overhead induced by~MPC is feasible even for large-scale deployments.
For example, one round of training~LeNet privately with security against semi-honest servers on the~MNIST dataset for~5 epochs with batch size~80 takes less than~5 minutes for a cluster with~10 clients, and global aggregation among~10 clusters takes less than~1 second.
Moreover, we find that the impact of~MPC-induced inaccuracies is less than~0.1\%, and models trained with~\FLname{} converge significantly faster than in plain~FL.

In~\tabref{tab:relatedworksconcise}, we summarize how~\FLname distinguishes itself from related works.
With respect to existing works on collaborative learning while also achieving global model privacy, we experimentally demonstrate superior scalability and performance.
For example, compared to the state-of-the-art framework~\Hercules~\cite{TDSC:XuHXZLHD23}, which is based on multi-party homomorphic encryption, we demonstrate convolutional neural network training with~20$\times$ more clients, 4$\times$ less communication, and~250$\times$ less storage overhead.

In summary, we provide the following contributions:

\begin{compactitemize}%
  \item We initiate a discussion to explore why novel approaches are necessary to jointly address concerns regarding security, privacy, and scalability in~FL.
  \item We propose a new generic~(hierarchical)~FL framework called \FLname{} that achieves global model privacy, supports resource-constrained mobile or edge devices, and significantly limits the attack surface for malicious clients to data poisoning.
  \item 
  We systematically evaluate the performance of~\FLname instantiations as well as its resilience against state-of-the-art poisoning attacks. Additionally, we provide a prototype implementation and evaluation of~\FLname on standard image classification tasks.
  \item 
  We experimentally compare~\FLname to state-of-the-art collaborative learning frameworks with global model privacy, demonstrating superior scalability and performance.
\end{compactitemize}
\section{Preliminaries and Related Work}
\label{sec:prelims}
This section provides a brief background on privacy-enhancing technologies like~MPC and~HE, and an introduction to~FL, along with a concise overview of related works.
We provide a comparative positioning in~\tabref{tab:relatedworksconcise} showing that no prior work simultaneously provides global model privacy and protection against malicious clients.

\myparagraph{Secure Multi-Party Computation~(MPC)}
\label{sec:related_work_mpc}
MPC~\cite{FOCS:Yao86,STOC:GolMicWig87} enables a set of mutually distrusting parties to evaluate a public function~$f()$ on their private data while preserving input data privacy. The corruption among the parties is often modelled via an~\emph{adversary} that may try to take control of the corrupted parties and coordinate their actions. There are various orthogonal aspects of \mbox{adversarial} corruption like honest vs.\ dishonest majority, semi-honest vs.\ malicious corruption, etc.~\cite{FTPS:EvansKR18,JACM:Lin20}. 
MPC is practically efficient when run among a small number of computing parties~\cite{NDSS:PatSur20,NeurIPS:KnottVHSIM21,ICML:Keller022,NDSS:KotiPRS22,PETS:CSWYCA25}.
In our prototype implementation\footnote{Our framework is adaptable to various settings and threat models, as discussed in~\secref{sec:framework-threat-model}. We built our implementation on CrypTen due to its maturity and ML-friendly design.} of~\FLname{}, we utilize~Meta's~CrypTen framework~\cite{NeurIPS:KnottVHSIM21}, which for efficiency reasons operates in a semi-honest two-party setting with a trusted third~\enquote{helper} party that generates correlated randomness~\cite{RiaziWTS0K18,CCSW:CCPS19,USENIX:PSSY21}.

\myparagraph{Homomorphic Encryption~(HE)}
\label{sec:related_work_he}
\makeparafit
HE schemes~\cite{Rivest1978,STOC:Gentry09} enable computation on encrypted data without the need for decryption. Additively homomorphic encryption~(AHE) is a widely used method that allows for the generation of a new ciphertext representing the sum of multiple plaintexts through operations on their corresponding ciphertexts~\cite{EUROCRYPT:paillier1999public}. In scenarios involving multiple parties, recent multi-party homomorphic encryption~(MHE) schemes can reduce the communication complexity of MPC, but have high computation overhead~\cite{PETS:mouchet2021multiparty}.  Furthermore, they are often secure only against semi-honest corruptions, whereas~\cite{EPRINT:PELTA23} is an exception. A comprehensive survey of various~HE schemes is given in~\cite{CSURV:carAUC18}.

\myparagraph{Differential Privacy~(DP)}
\label{sec:related_work_dp}
The concept of~DP~\cite{TTC:DworkR14} is based on the idea of adding noise to data in order to reduce information leakage when sharing it, while still allowing for meaningful computations to be carried out on the data. Though DP techniques offer some protection against attacks in federated learning, they also reduce the utility of the data and the resulting ML model. Additionally, achieving robust and accurate models necessitates significant privacy budgets, leaving the actual level of privacy achieved in practice uncertain~\cite{ACCESS:OA22}. We refer to~\cite{PETS:DP20} for more details on DP.

\myparagraph{Federated Learning~(FL)}
\label{sec:related_work_fl}
Unlike conventional~PPML techniques, FL~\cite{ARXIV:konevcny2016federated,AISTATS:McMahanMRHA17} enables the training of an ML model on distributed data by allowing each device to train the model locally using its own data. In each iteration, these local models are sent to a central server (also called an aggregator), where they are \emph{aggregated} to form a global model. At a high level, an~FL scheme iterates through the following steps:
\begin{myitemize}
    \item  The global server~$\server$ sends the current global model~$\FLmodel{}{t}$ to a selected subset of~$n$ out of~$N$ clients.
    \item Each selected client~$C^i$, $i \in [n]$ utilizes its own local training data~$D^i$ for~$E$ epochs to fine-tune the global model and obtains an updated local model~$w_{t+1}^{i}$:
    \[
      w_{t+1}^{i} \leftarrow W_{t}-\eta_{C^i} \frac{\partial L (W_{t},B_{i,e})}{\partial W_{t}},
    \]
    where~$L$ is a loss function, $\eta_{C^i}$ is the clients' learning rate, and~$B_{i,e}\subseteq D^i$ is a batch drawn from~$D^i$ in epoch~$e$, where~$e \in [E]$. The local model updates~$w_{t+1}^{i}$ are then sent back to~$\server$.
    \item $\server$ employs an aggregation rule~$f_{\textit{agg}}$ to combine the received local model updates~$w_{t+1}^{i}$, resulting in a global model~$W_{t+1}$, which will serve as the starting point for the next iteration:
    \[
      W_{t+1} \leftarrow W_{t}-\eta_\mathsf{S} \cdot f_{\textit{agg}}(w_{t+1}^{1},\ldots,w_{t+1}^{n}),
    \]
    where~$\eta_\mathsf{S}$ is the server's learning rate.
\end{myitemize}

The above procedure is repeated until a predefined stopping criterion is satisfied, e.g., a specified number of training iterations or a specific level of accuracy.

\begin{sidewaystable}
    \centering
    \captionsetup{font=small}
    \caption{Comparison of \FLnameTitle{} and previous works. Notations: $\server$ -- Aggregation Server(s), $\client$ -- Client, GM -- Global Model, LM -- Local Model, AHE/MHE -- Additively/Multi-Party Homomorphic Encryption, ZKP -- Zero-Knowledge Proof, MPC -- Secure Multi-party Computation, DP -- Differential Privacy. In terms of privacy, we focus on the protection of the global model on both server and client side as well as the protection of individual client models against a curious or malicious server. Given the extensive literature, this comparison focuses on one representative work per category.}
    \label{tab:relatedworksconcise}
    \begin{small}
    \begin{threeparttable}
        \begin{tabular}{lccccccccc} 
        \toprule
        \multirow{2}{*}{Categories} 
        & \multicolumn{1}{c}{\multirow{2}{*}{\makecell{Representative\\Work(s)}}}
        & \multicolumn{1}{c}{\multirow{2}{*}{\makecell{Privacy\\Method}}}
        & \multicolumn{2}{c}{Privacy ($\server$)} 
        & Privacy ($\client$) 
        & \multirow{2}{*}{Defense}  
        & \multirow{2}{*}{\makecell{Cross\\Device}} 
        & \multirow{2}{*}{\makecell{No Client\\Interaction}} 
        & \multirow{2}{*}{\makecell{Dropout\\Handling}}
        \\ \cmidrule{4-6}
        & & & GM & LM & GM & &  & &
        \\ \midrule
        Aggregation~(Plain)
        & {\footnotesize \cite{AISTATS:McMahanMRHA17}} 
        &--&\xmark&\xmark&\xmark&\xmark&\cmark&\cmark&\cmark\\ 
        \midrule
        \multirow{2}{*}{Aggregation~(Robust)} 
        & {\footnotesize \cite{NIPS:BlanchardMGS17}} 
        &--&\xmark&\xmark&\xmark&\cmark&\xmark&\cmark&\cmark\\  
        &{\footnotesize \cite{NDSS:CaoF0G21,AISTATS:ZWPWJSJ23}}
        &--&\xmark&\xmark&\xmark&\cmark&\cmark&\cmark&\cmark\\ 
        \midrule
        \multirow{4}{*}{\makecell[l]{Secure Aggregation\\($\singleserverSA$)}} 
        & {\footnotesize \cite{CCS:BIKMMP17}}
        & Masking &\xmark & \cmark &\xmark &\xmark &\xmark &\xmark &\cmark\\ 
        &{\footnotesize \cite{CCSW:MandalG19}}
        &AHE&\xmark&\cmark&\cmark&\xmark&\xmark&\cmark&\cmark\\
        &{\footnotesize \cite{ICCV:GJSZ0K23}}
        & Masking+ZKP&\xmark&\cmark&\xmark&\cmark&\xmark&\xmark&\cmark\\
        &{\footnotesize \cite{NIPS:MYP24,AAAI:LBGPHD25}}
        & Masking&\xmark&\cmark&\xmark&\cmark&\xmark&\xmark&\cmark\\
        \midrule
        
        \multirow{6}{*}{\makecell[l]{Secure Aggregation\\($\multiserverSA$)}}
        &{\footnotesize \cite{DLS:FereidooniMMMMN21}}
        &MPC&\cmark&\cmark&\xmark&\xmark&\cmark&\cmark&\cmark\\
        &{\footnotesize \cite{PETS:FTPSSB21}}\tnote{$\ast$}
        &MHE&\cmark&\cmark&\cmark&\xmark&\xmark&\xmark&\xmark\\
        &{\footnotesize \cite{NDSS:SavPTFBSH21,TDSC:XuHXZLHD23}\tnote{$\ast$}}
        &MHE&\cmark&\cmark&\cmark&\xmark&\xmark&\xmark&\xmark\\
        & {\footnotesize \cite{USENIX:NguyenRCYMFMMMZ22}}
        &MPC&\cmark&\cmark&\xmark&\cmark&\xmark&\cmark&\cmark\\ 
        & {\footnotesize \cite{SP:RatheeSWP23}}
        &MPC&\cmark&\cmark&\xmark&\xmark\tnote{$\ddagger$}&\cmark&\cmark&\cmark\\   
        & {\footnotesize \cite{TCHES:TXLZ25}}
        &MPC&\cmark&\cmark&\xmark&\xmark\tnote{$\ddagger\star$}&\cmark&\cmark&\cmark\\   
        \midrule
        \multirow{5}{*}{\makecell[l]{Hierarchical FL\\(HFL)}} 
        & {\footnotesize \cite{MLSys:BonawitzEGHIIKK19}}
        & Masking &\xmark&\cmark&\xmark&\xmark&\cmark&\xmark&\cmark\\ 
        & {\footnotesize \cite{INFOCOM:WangXLHQZ21}} 
        &--&\xmark&\xmark&\xmark&\xmark&\cmark&\cmark&\cmark\\ 
        & {\footnotesize \cite{IJCAI:Yang21}}
        &DP&\xmark &\cmark &\xmark &\xmark&\cmark&\cmark&\cmark\\ 
        & {\footnotesize \cite{NIPS:FHC0B24}}
        &--&\xmark&\xmark&\xmark&\xmark&\cmark&\cmark&\cmark\\ 
        & {\footnotesize \cite{TIFS:WHZLWG25}} 
        & Masking &\xmark&\cmark&\xmark&\cmark&\cmark&\cmark&\cmark\\ 
        \midrule
        Worldwide FL (\FLnameTitle) 
        & \textbf{This work} 
        & MPC\tnote{$\dag$} 
        &\cmark&\cmark&\cmark&\cmark&\cmark&\cmark&\cmark\\ 
        \bottomrule
        \end{tabular}
        {\footnotesize
        \begin{tablenotes}
            \item{$\ast$} Represents collaborative learning in the $N$-party setting, rather than the Federated Learning (FL) setting. SPINDLE~\cite{PETS:FTPSSB21} supports only generalized linear models and lacks protocols for training complex models like NNs; POSEIDON~\cite{NDSS:SavPTFBSH21} and HERCULES~\cite{TDSC:XuHXZLHD23} use activation function approximations, reducing final accuracy.
            \item{$\ddagger$} Limited to defenses that operate independently on each client's gradient; excludes aggregation methods like trimmed mean or median~\cite{ICML:YinCRB18}.
            \item{$\star$} Cannot detect malformed gradients that lie within the norm bound (e.g., shrunk gradients).
            \item{$\dag$} \FLname{} is a generic design that can be implemented with various secure computation protocols, such as MPC or HE.
        \end{tablenotes}
        }
    \end{threeparttable}
    \end{small}
\end{sidewaystable}

\myparagraph{Secure Aggregation~($\SA$)}
\label{sec:secure_aggregation_related}
Standard FL's privacy issues~\cite{NIPS:GeipingBD020,SP:ZSEEAB24,ICLR:WBSCZXS24,USENIX:DHWSGRC25} led to single-server secure aggregation protocols~($\singleserverSA$)~\cite{CCS:BIKMMP17,CCS:BellBGL020,PETS:MOJC23} that use masking techniques or HE~\cite{cryptologia:Rubin96a,TIFS:PhongAHWM18,CCSW:MandalG19}. However, complete privacy for the global model remains elusive, posing risks if clients and aggregators collude. Moreover, recent research has shown that a single malicious server can reconstruct individual training data points from users' local models even when using secure aggregation protocols~\cite{CCS:PFA22,ARXIV:BDSSSP23,SP:ZSEEAB24,INFOCOM:WCHPDCR24}. Such vulnerabilities against colluding parties are addressed by multi-server $\SA$ ($\multiserverSA$)~\cite{DLS:FereidooniMMMMN21,ESORICS:DongCLWZ21,USENIX:NguyenRCYMFMMMZ22,SP:RatheeSWP23,TCHES:TXLZ25} and techniques like multi-party homomorphic encryption~(MHE)~\cite{PETS:FTPSSB21,NDSS:SavPTFBSH21,TDSC:XuHXZLHD23}, albeit with scalability challenges. Works like~\cite{ESORICS:DongCLWZ21} and~\cite{USENIX:NguyenRCYMFMMMZ22} combine~HE and~MPC techniques to achieve private and robust~FL, albeit at the cost of considerable computation and communication overhead in cross-device scenarios comprising hundreds to thousands of clients~\cite{SP:SHKR22}. In~\FLname, we employ a multi-server aggregation scheme, and mitigate scalability issues of existing works, especially for a heterogeneous cross-device setting.

\myparagraph{Collaborative Learning with Global Model Privacy}
\label{sec:collaborative-learning-model-privacy}
Several prior works focus explicitly on preserving the privacy of the global model during collaborative learning. 
These approaches primarily rely on~HE, particularly multi-party homomorphic encryption~(MHE)~\cite{PETS:mouchet2021multiparty}, to securely aggregate model updates while maintaining privacy across multiple clients.
MHE enables distributed and fully decentralized training through direct interactions among data owners in settings where non-colluding servers are unavailable.
\Spindle~\cite{PETS:FTPSSB21} employs~MHE to support secure distributed training, specifically for logistic regression models.
However, \Spindle is limited to linear models, which restricts its applicability to more complex architectures.
\Poseidon~\cite{NDSS:SavPTFBSH21} extends~MHE to~3-layer neural networks by leveraging~multi-party~CKKS~\cite{PETS:mouchet2021multiparty}.
In this framework, multiple entities collaboratively generate a shared secret key, enabling encrypted computations under a common public key.
To maintain computational efficiency, \Poseidon replaces non-polynomial functions, such as the commonly used~ReLU activation function, with polynomial approximations.
However, these polynomial approximations introduce accuracy trade-offs that may significantly impact model performance.
To address these computational inefficiencies, \Hercules~\cite{TDSC:XuHXZLHD23} was proposed as an enhancement of~\Poseidon.
\Hercules introduces optimized parallel homomorphic operations, including~SIMD-enabled matrix computations and improved polynomial approximations for~RELU.
While~\Hercules offers higher computational performance than~\Poseidon, it continues to suffer from accuracy degradation due to the use of polynomial approximations for non-linear layers.

Despite these advancements, all three MHE-based frameworks -- \Spindle~\cite{PETS:FTPSSB21}, \Poseidon~\cite{NDSS:SavPTFBSH21}, and~\Hercules~\cite{TDSC:XuHXZLHD23} -- exhibit notable limitations.
The reliance on polynomial approximations for non-linear functions, particularly~ReLU, inherently reduces model accuracy~\cite{CCS:RRKCGRS20}.
Moreover, these frameworks lack dedicated robustness mechanisms against poisoning attacks, without which a single malicious participant can substantially degrade the global model accuracy~\cite{USENIX:FangCJG20}.
Scalability also remains a critical limitation: the experiments in these works typically involve no more than~50 parties, raising concerns about their feasibility in larger, real-world collaborative scenarios.
Additionally, these approaches require interactive key setup among clients for encrypting each model and cooperative decryption of the final aggregated model, which introduces further complexity and communication overhead.

Another significant bottleneck in~HE-based frameworks is their high storage complexity.
For example, training on the~MNIST dataset using a lightweight~3-layer fully connected model, \Poseidon suffers from an exceptionally high local storage burden -- up to~314GB per client.
This results from its packing encryption method, which replicates encrypted values and inserts zero-padding to match layer sizes, depending on the number of neurons in each hidden layer.
Such replication leads to multiple ciphertext copies and excessive padding, rendering the approach inefficient for client-side deployment and impractical for large-scale systems.
While~\Hercules improves computational efficiency, it still incurs a~4.9GB local storage overhead per client -- potentially prohibitive for resource-constrained devices.
In contrast, \FLname{} avoids this overhead entirely by eliminating ciphertext storage on the client side: clients keep their training data in plaintext, while servers hold secret shares of both training data and model parameters.
The size of secret shares might not always exactly match that of the plaintext; they are typically larger by a small constant factor~(e.g., 2$\times$-3$\times$), depending on the secret-sharing scheme and number of servers. 
This design makes our framework not only faster, but also significantly lighter in terms of storage requirements, particularly benefiting resource-constrained clients such as~IoT devices.

\myparagraph{Hierarchical Federated Learning~(HFL)}
\label{sec:related_work_hfl}
Another approach for~FL, called hierarchical~FL~(HFL) addresses scalability and heterogeneity issues in real-world systems~\cite{MLSys:BonawitzEGHIIKK19,AAAI:WangWCJ22}.
In~HFL, aggregation occurs at multiple levels, forming a hierarchical structure in which the aggregated values from one level serve as inputs for the above level.
This procedure eventually reaches the top level, where the final model is aggregated~\cite{ICDCS:DengLRZZZY21,INFOCOM:WangXLHQZ21}.
Most of the works in the~HFL setting have primarily focused on improving scalability and communication~\cite{NIPS:FHC0B24,TIFS:WHZLWG25}.
An exception is the work by~\cite{IJCAI:Yang21}, which proposed methods to ensure the privacy of individual client updates in~HFL, allowing for secure aggregation. However, their approach does not address global model privacy or provide robustness against malicious users, which we provide in~\FLname{}.

\myparagraph{Poisoning \& Robust Aggregation} 
\label{sec:related_work_poisoning}
One potential drawback of~FL over~PPML techniques is that~FL increases the attack surface for malicious actors: in~FL, each user trains their own model locally, allowing them to manipulate their model~\cite{ICML:BCMC19,NDSS:ShejwalkarH21}. To counter these~\emph{poisoning} attacks, various robust aggregation methods have been proposed, in which local models are analyzed and appropriate measures such as outlier removal or norm scaling are applied~\cite{ICML:YinCRB18,AAAI:li2019rsa,NDSS:CaoF0G21,SATML:BMPSTVWYY24}. See~\cite{SP:SHKR22,CSUR:GGP24,CSUR:ZBACCDLLNR25} for a comprehensive overview of different robust aggregations.

\myparagraph{Model Extraction Attacks} 
\label{sec:related_work_extraction}
In model extraction attacks, an adversary leverages black-box access to a model’s outputs to deduce its structure without prior knowledge~\cite{ASIACCS:LiangPLW24,CSUR:ZBACCDLLNR25}. While existing privacy-preserving FL frameworks using MPC or HE protect data privacy, they do not prevent such attacks, as the adversary still has access to model outputs. We consider model extraction prevention orthogonal to private inference in FL and suggest incorporating methods like rate limiting and differential privacy~\cite{KDD:ZLDGZD25} for future work.
\section{The \FLnameTitle{} Framework}
\label{sec:framework}
We now present the details of our~\FLname{} framework, which addresses multiple key requirements in~FL: global model privacy, scalability, support for resource-constrained~(mobile/edge) devices, reduction of attack surface, ability to defend against data poisoning, and high levels of user engagement. Our framework captures various proposed architectures for~FL in a unified abstraction. We illustrate our framework in~\figref{fig:architecture} and detail the underlying training algorithm in~\algref{alg:algorithm-workflow}.
The notations used in~\FLname{} are listed in~\tabref{tab:notation}.

\begin{table}[htb!]
    \centering
    \captionsetup{font=small}
    \caption{Notations used in our \FLnameTitle{} framework.}
    \small
    \label{tab:notation}
    \begin{small}
    \begin{tabular}{rl}
       \toprule
       Notation  & Description\\
       \midrule
       $\globalserverset$ &
       Set of Global MPC Servers.\\
       $\clusterserverset{i}$ &
       Set of MPC Cluster Servers in the $i$-th cluster.\\
       $\clientFLsetTotal{i}$ &
       Set of Clients in the $i$-th cluster.\\
       $\clientFLset{i}$ &
       Selected Clients in the $i$-th cluster.\\
       $\sizeFL{s}$    &
       Size of the set $s \in \{\globalserverset, \clusterserverset{i}, \clientFLset{j}\}$.\\
       $\numclusters$  &
       Total number of clusters.\\
       $\WWFLmodel{t}$ & 
       Global model available at round $t$.\\
       $\globalserver{i}$ &  
       Layer I MPC Global Server; $\globalserver{i} \in \globalserverset$.\\
       $\clusterserver{i}{j}$ & 
       \makecell[l]{Layer II MPC Cluster Server; $\clusterserver{i}{j} \in \clusterserverset{i}$. Here, $i \in [\numclusters], j \in [\clusterserversize{i}]$}\\
       $\clientFL{i}{j}$ & 
       Layer III Client; $\clientFL{i}{j} \in \clientFLset{i}$, $i \in [\numclusters], j \in [\clusterclientsize{i}]$\\
       $\MPCSharing{\cdot}{s}$ & 
       Secret sharing semantics for $s \in \{\globalserverset, \clusterserverset{j}\}$.\\
       \bottomrule
    \end{tabular}
    \end{small}
    \vspace{-2mm}
\end{table}


\begin{figure*}[t]
    \centering
    \resizebox{\textwidth}{!}{
    \input{figures/architecture.tex}
    }
    \vspace{-3mm}
    \captionsetup{font=small}
    \caption{Three-layer architecture in \FLnameTitle{} for federated training of a machine learning model. The number of servers and clients shown is illustrative and not fixed. See~\secref{sec:system-flow} for further details.}
    \label{fig:architecture}
    \vspace{-5mm}
\end{figure*}

\subsection{\FLnameTitle{} Architecture}
\label{sec:framework-layers}
Our~\FLname{} framework is based on a three-layer\footnote{Additional MPC Cluster layers can be easily incorporated depending on the scale of deployment.} architecture and extends the established hierarchical~FL~(HFL) paradigm~\cite{MLSys:BonawitzEGHIIKK19,INFOCOM:WangXLHQZ21,IJCAI:Yang21}.
In~HFL, clients are initially organized into clusters, and their local models are aggregated at the cluster level;
these cluster-level models are then further aggregated globally, resulting in an additional layer of aggregation.
We adopt the hierarchical approach in~\FLname{} for two main reasons: 1) to efficiently realize our distinct model training approach for global model privacy, and 2) to facilitate large-scale deployments across heterogeneous clients. However, there is one key difference: at the cluster level, \FLname{} does not perform aggregation of models but private training on collective client data. This differs from both traditional and hierarchical~FL since the data is not kept solely on the client devices. Nevertheless, \FLname{} ensures that the data remains within the trusted domain of the client, allowing for accurate modeling of trust relationships between clients in real-world situations, such as trust among members of the same region or country~\cite{CN:MartiG06}. Furthermore, this type of hierarchy is ubiquitous in real-world scenarios such as~P2P gaming, organizations, and network infrastructures~\cite{INFOCOM:SARK02,CSUR:HV19,ICLR:LSPJ20}.

We provide details for each layer in~\FLname next.
We focus on the sequence of operations performed by the entities in our architecture~(cf.~\figref{fig:architecture}) for a training over~$T$ iterations while considering necessary~MPC protocols and setup requirements. To make our design generic, we use the $\funcMPC$-hybrid model, a standard way to abstract MPC protocols. Concrete instantiations are given in~\tabref{tab:mpc-functionalities} in~\secref{sec:mpc-functionalities}. 

\myparagraph{Layer III: Clients}
\label{sec:framework-layer-III}
This layer is composed of~$\numclusters$ distinct sets of clients~$\clientFLsetTotal{i}$~(with~$i \in [\numclusters]$), called~\emph{clusters}, which are formed based on specific criteria of the application~(e.g., European Union~(EU) member states for the~EU smart metering scheme~\cite{CHAPTER:CK13,REPORT:EUSmartMeter}).
Like standard~FL, only a random subset of clients~$\clientFLset{i} \subseteq \clientFLsetTotal{i}$ will be selected by the training algorithm in an iteration~$t \in [1,T]$.
This subset selection is performed by the MPC clusters in Layer~II using the $\MPCSample$ algorithm, ensuring that the selection process adheres to the desired security guarantees.

\begin{table}[hb!]
    \captionsetup{font=small}
    \caption{MPC functionalities used in \FLnameTitle{}.}
    \label{tab:mpc-functionalities}
    \centering
    \begin{small}
    \begin{tabular}{rrp{8.5cm}}
       \toprule
       Algorithm & Input(s)  & Description\\
       \midrule
       $\MPCShare$ & $\MPCData{}{}, \MPCTarget$ &
       Generates secret shares of data $\MPCData{}{}$ as per target $\MPCTarget$'s sharing semantics~($\MPCSharing{\MPCData{}{}}{\MPCTarget}$) with~$\MPCTarget\in\{\clientFLset{},\clusterserverset,\globalserverset\}$.\\
       $\MPCReshare$ & $\MPCSharing{\MPCData{}{}}{\MPCSource}, \MPCTarget$ &
       Converts secret shares of data $\MPCSharing{\MPCData{}{}}{}$~from~source $\MPCSource$'s sharing semantics to target $\MPCTarget$.\\
       $\MPCAgg$ & $\{ \MPCSharing{\MPCModel{}{j}}{} \}$& Performs secure aggregation over $j$ secret-shared ML models.\\
       $\MPCTrain$ & $\MPCSharing{\MPCModel{}{}}{}, \MPCSharing{\MPCData{}{}}{}$ & Performs PPML Training on ML model $\MPCModel{}{}$ using the secret-shared data $\MPCSharing{\MPCData{}{}}{}$.\\
       $\MPCInf$ & $\MPCSharing{\MPCModel{}{}}{}, \MPCSharing{\MPCQuery{}{}}{}$ & Performs PPML Inference on secret-shared ML model $\MPCSharing{\MPCModel{}{}}{}$ using the secret-shared query $\MPCSharing{\MPCQuery{}{}}{}$.\\
       $\MPCReveal$ & $\MPCSharing{\MPCData{}{}}{\MPCSource}, \MPCTarget$ &
       Reconstructs secret-shared data $\MPCSharing{\MPCData{}{}}{}$ towards members in target set $\MPCTarget$.\\
       $\MPCSample$ & $\clientFLsetTotal{}$ &
       Selects a subset of clients $\clientFLset{}$ from $\clientFLsetTotal{}$.\\
       $\MPCTrimmedMeanList$ & $\FLmodelSet{}, \TRParam$ & Performs Trimmed Mean and returns $2\TRParam$ outlier values (top and bottom $\TRParam$) for each index position in elements of $\FLmodelSet{}$.\\
       $\MPCTopKHitter$ & $\myMPCSet{U}, \Gamma$ & Returns list of~$\Gamma$ values that occur most frequently in~$\myMPCSet{U}$.\\
       \bottomrule
    \end{tabular}
    \end{small}
\end{table}

\makeparafit
During iteration~$t$, client~$\clientFL{i}{j} \in \clientFLset{i}$~(with~$j \in [\clusterclientsize{i}]$) holding data~$\WWFLData{t}{\clientFL{i}{j}}$ uses the~$\MPCShare$ protocol~(cf. \tabref{tab:mpc-functionalities}) to securely distribute its data to a set of MPC cluster servers~$\clusterserverset{i}$. $\clusterserverset{i}$ are a representative group of high-performance servers that are trusted by their clients not to collude entirely, though some level of collusion among a subset of servers may occur.
\FLname{} allows clients to share their input and then leave at any time.
They can also rejoin the system later and provide additional data in the next iteration they get selected.
Hence, the clusters~$\clientFLsetTotal{i}$ are dynamic and change with each iteration.

Our method differs from the standard concept of \enquote{data residing at the clients} in~FL, but we expect it to not negatively impact user engagement as data remains within the users' trust zone. 
Additionally, the reduced computational load allows resource-constrained devices to train complex models and eliminates the need for a shared-key setup among clients, simplifying dropout handling.

\myparagraph{Layer II: MPC Clusters}
\label{sec:framework-layer-II}
The second layer consists of~$\numclusters$ sets of distributed training servers~$\clusterserverset{i}$~(with~$i \in \numclusters$), called~\emph{MPC clusters}, with each~$\clusterserverset{i}$ corresponding to cluster~$\clientFLsetTotal{i}$ in~Layer~III.
In iteration~$t$, Layer~I servers~(denoted by $\globalserverset$) initiate~ML training by sharing the current global model~$\WWFLmodel{t-1}$ among the servers in~$\clusterserverset{i}$. As will be discussed in~\secref{sec:framework-layer-I}, $\WWFLmodel{t-1}$ is also in a secret-shared form among~$\globalserverset$, represented by~$\MPCSharing{\WWFLmodel{t-1}}{\globalserverset}$. To account for varying availability and trustworthiness of servers across regions, MPC clusters in~\FLname{} may use different~MPC configurations and differ, e.g., in their corruption threshold and security model~\cite{FTPS:EvansKR18}. Therefore, $\globalserverset$ uses the~$\MPCReshare$ protocol~(cf.~\tabref{tab:mpc-functionalities}) to convert the secret shares of~$\MPCSharing{\WWFLmodel{t-1}}{\globalserverset}$ to those of~$\clusterserverset{i}$, i.e., $\MPCSharing{\WWFLmodel{t-1}}{\clusterserverset{i}}$.

Given~$\MPCSharing{\WWFLmodel{t-1}}{\clusterserverset{i}}$, the servers in~$\clusterserverset{i}$ use the~$\MPCTrain$ protocol (cf.~\tabref{tab:mpc-functionalities}) to run MPC-based~PPML for private~ML training~\cite{NeurIPS:KnottVHSIM21,ICML:Keller022} on the cumulative secret-shared data from all clients in the cluster~$\clientFLset{i}$, denoted by~$\MPCSharing{\WWFLData{t}{}}{\clusterserverset{i}}$.
This data may include data from the same cluster in the previous iteration.
Furthermore, by utilizing a larger pool of training data, we can leverage the known benefits of batching, resulting in faster convergence~\cite{ARXIV:GDGNWKTJH17,SIAM:BCN18}.
After completing training, the servers in~$\clusterserverset{i}$ run~$\MPCReshare$ to secret-share the updated model with the~Layer~I servers, i.e., $\MPCSharing{\FLmodel{i}{t}}{\globalserverset}$. 

To preserve the system's integrity, the servers for each~MPC cluster must be chosen with care to ensure clients are willing to share their data among the servers and that not all servers are colluding.
One option is to build non-profit partnerships, such as in the~MOC alliance~\cite{CLOUDNET:ZinkICSKHDDLH21}, where organizations with mutual distrust can securely co-locate servers in the same data center with high-speed network connections.
Alternatively, trusted entities like government organizations with limited infrastructure can host their servers in confidential cloud computing environments~\cite{CACM:RCFCD21}.
If a client lacks trust in the~\FLname{} servers in its current cluster, the client can propose a new cluster with a partner that provides a non-colluding server.

\myparagraph{Layer I: Global Servers}
\label{sec:framework-layer-I}
The top layer consists of a set of~MPC servers~$\globalserverset$, named~\emph{Global Servers}, that~\emph{securely} aggregate trained models from all the~MPC clusters in~Layer~II, similarly to a standard~FL scheme with a distributed aggregator~\cite{DLS:FereidooniMMMMN21,SP:RatheeSWP23,DLS:GMSSWY23,SATML:BMPSTVWYY24,TCHES:TXLZ25}.
Given the locally trained models~$\FLmodel{i}{t}$ for~$i \in [\numclusters]$, the servers in~$\globalserverset$ execute the secure aggregation protocol~$\MPCAgg$~~\cite{PETS:MOJC23} to compute the updated global model in secret-shared form, i.e.,~$\MPCSharing{\FLmodel{}{t}}{\globalserverset}$.
The global servers in~$\globalserverset$ use the~$\MPCReshare$ protocol to distribute the aggregated model~$\FLmodel{}{t}$ to each of the~Layer~II clusters~$\clusterserverset{i}$ for the next iteration~$t+1$.

The distributed aggregator model, as outlined in~\secref{sec:intro}, circumvents the possibility of several privacy attacks that were recently presented for single-server aggregators even when relying on maliciously secure aggregation protocols~\cite{WEB:BDSSSP22,ICLR:FowlGCGG22,ICML:WenGFGG22,AAAI:SAGJA23,EUROSP:BDSSSP23,SP:ZSEEAB24}; for example, by excluding certain inputs from the computation and thus influencing the \emph{correctness}~(but not the security) of the protocol, a malicious single server can mount \emph{privacy} attacks that allow to infer information about particular participants and their training data.
In contrast, in our distributed aggregator setting, even when having only semi-honest protocol security, such attacks are detectable as long as at least one of the servers behaves honestly.

\begin{algorithm}[htb!]
   \begin{small}
   \caption{\FLTrain: Training in~\FLnameTitle{}}
   \label{alg:algorithm-workflow}
    \begin{algorithmic}[1]
       \ACTORS $\globalserverset, \clusterserverset{}, \clientFLset{}$ 
       \COMMENT{$\clusterserverset{} = \bigcup_i \clusterserverset{i}$, $\clientFLset{} = \bigcup_i \clientFLset{i}$; $i \in \numclusters$}
       \INPUT $\WWFLmodel{0}$, $\{\WWFLData{}{\clientFL{}{}}\}_{\clientFL{}{}\in\clientFLset{}}$ \COMMENT{$\WWFLmodel{0}$ -- initial model, $\WWFLData{}{\clientFL{}{}}$ -- client $\clientFL{}{}$'s data}
       \OUTPUT $\MPCSharing{\WWFLmodel{T}}{}$ \COMMENT{$\WWFLmodel{T}$ -- global model after $T$ iterations} 
       \STATE {\bfseries initialize:} $\MPCSharing{\WWFLmodel{0}}{\globalserverset} \leftarrow \MPCShare(\WWFLmodel{0}, \globalserverset)$ \label{line:fltrain-step-a}
       \FOR{each training iteration $t \in [1,T]$}
            \FORALL[in parallel]{$i \in \numclusters$}
                \STATE{$\MPCSharing{\WWFLmodel{t-1}}{\clusterserverset{i}} \leftarrow \globalserverset.\MPCReshare(\MPCSharing{\WWFLmodel{t-1}}{\globalserverset}, \clusterserverset{i})$\COMMENT{global servers $\globalserverset$ reshare $\WWFLmodel{t-1}$ with cluster servers}} \label{line:fltrain-step-b}
                \STATE{$\clientFLset{i} \leftarrow \clusterserverset{i}.\MPCSample(\clientFLsetTotal{i}, t)$\COMMENT{$\clientFLsetTotal{i}$ -- total clients in $i$-th cluster}}
                \FORALL[in parallel]{$j \in [\clusterclientsize{i}]$} 
                    \STATE{$\MPCSharing{\WWFLData{t}{\clientFL{i}{j}}}{\clusterserverset{i}} \leftarrow {\clientFL{i}{j}}.\MPCShare(\WWFLData{t}{\clientFL{i}{j}}, \clusterserverset{i})$\COMMENT{$\WWFLData{t}{\clientFL{i}{j}}$ -- client $\clientFL{i}{j}$'s data in iteration $t$}}
                \ENDFOR
                \STATE{$\MPCSharing{\WWFLData{t}{}}{\clusterserverset{i}} \leftarrow  \bigcup\limits_{j \in [\clusterclientsize{i}]} \MPCSharing{\WWFLData{t}{\clientFL{i}{j}}}{\clusterserverset{i}} \bigcup \MPCSharing{\WWFLData{t-1}{}}{\clusterserverset{i}}$\COMMENT{$\WWFLData{0}{}=\emptyset$}}
                \STATE{$\MPCSharing{\FLmodel{i}{t}}{\clusterserverset{i}} \leftarrow {\clusterserverset{i}}.\MPCTrain(\MPCSharing{\WWFLmodel{t-1}}{\clusterserverset{i}}, \MPCSharing{\WWFLData{t}{}}{\clusterserverset{i}})$} \label{line:fltrain-step-c}
                \STATE{$\MPCSharing{\FLmodel{i}{t}}{\globalserverset} \leftarrow {\clusterserverset{i}}.\MPCReshare(\MPCSharing{\FLmodel{i}{t}}{\clusterserverset{i}}, \globalserverset)$} \label{line:fltrain-step-d}
            \ENDFOR
            \STATE{$\MPCSharing{\WWFLmodel{t}}{\globalserverset} \leftarrow {\globalserverset}.\MPCAgg(\{\MPCSharing{\FLmodel{i}{t}}{\globalserverset}\}_{i \in [\numclusters]})$} \label{line:fltrain-step-e}
       \ENDFOR
    \end{algorithmic}
    \end{small}
\end{algorithm}

\myparagraph{\FLnameTitle{} -- The Complete Picture}
\label{sec:framework-workflow}
\algref{alg:algorithm-workflow}~(\FLTrain) provides the training algorithm in~\FLname. We detail the necessary~MPC functionalities along with the~\FLname{} training functionality $\funcFLname$ (\algref{alg:algorithm-ideal-functionality}) in~\secref{sec:framework-threat-model}.
Although~\FLname{} has a 3-layer architecture, it can easily accommodate more levels of hierarchy depending on the size of the deployment. For this, additional layers of~MPC clusters can be added between layers~I~and~II, with the clusters performing secure aggregation instead of~PPML training.

Existing schemes for global model privacy, such as~\cite{SP:RatheeSWP23} and~\cite{TCHES:TXLZ25}, only protect the model from either clients or aggregator servers, leaving the possibility of collusion especially in a cross-device setting. \FLname{} addresses this issue by keeping the global model in secret-shared form, ensuring that no single entity or group of colluding entities~(up to an allowed corruption threshold) can access the model. This provides a stronger notion of privacy and also protects against unauthorized use or misuse, such as a client disclosing the trained model to another organization for further training or commercial use. For scenarios demanding disclosure of the global model to authorized entities, such as consortium members, the global servers can utilize the $\MPCReveal$ protocol.

In~\secref{sec:system-flow}, we provide a detailed explanation of the $\FLTrain$ algorithm for a scenario involving three Layer~II MPC clusters, each using a different protocol: two-party~\cite{NDSS:Demmler0Z15}, three-party~\cite{CCS:AFLNO16,CCS:MohRin18}, and four-party~\cite{NDSS:KotiPRS22}. The global servers are instantiated using a two-party protocol with a helper, based on CrypTen~\cite{NeurIPS:KnottVHSIM21}.

\subsection{MPC Functionalities in~\FLnameTitle}
\label{sec:mpc-functionalities}
The~MPC functionalities utilized in \FLname{} are listed in~\tabref{tab:mpc-functionalities} and discussed in the following.
While~$\MPCShare$ and~$\MPCReshare$ are used for generating the secret shares as per the underlying~MPC semantics, $\MPCReveal$ is used to reconstruct the secret towards a designated party.
The~$\MPCTrain$ and~$\MPCInf$ functionalities correspond to~PPML training and inference protocols, respectively.
Similarly, $\MPCAgg$ denotes the secure aggregation functionality in~FL~(cf.~\secref{sec:secure_aggregation_related}).
In~\FLname{}, these functionalities are realized using~Meta's~CrypTen framework, in which two semi-honest~MPC servers carry out the computation with the help of a trusted third party that provides correlated randomness~\cite{CRYPTO:DamgardPSZ12,NDSS:Demmler0Z15,RiaziWTS0K18,NeurIPS:KnottVHSIM21}. 
However, \FLname{} is not bound to any specific~MPC setting and each of the MPC clusters as well as the global servers could be instantiated independently using any~MPC protocol, as detailed further in \secref{sec:framework-threat-model} and \secref{sec:framework-abstraction}. 

\myparagraph{$\funcReshare$ Ideal Functionality}
\label{sec:ideal-functionality-reshare}
Given the hybrid nature of~\FLname{}, the security of the framework heavily relies on the correct instantiation of the ideal functionality for the~$\MPCReshare$ algorithm, denoted as~$\funcReshare$.
This ideal functionality is presented in~\algref{alg:algorithm-ideal-functionality-reshare}.

\begin{algorithm}[htb!]
   \begin{small}
   \caption{$\funcReshare$: Ideal Functionality for~\MPCReshare}
   \label{alg:algorithm-ideal-functionality-reshare}
    \begin{algorithmic}[1]
       \INPUT $\partyset: \MPCSharing{\MPCData{}{}}{\MPCSource}$, $\partyset': \bot$ 
       \OUTPUT $\partyset: \bot$, $\partyset': \MPCSharing{\MPCData{}{}}{\MPCTarget}$ 
       \STATE {\bfseries secret reconstruction:} $\funcReshare$ receives $\MPCSharing{\MPCData{}{}}{\MPCSource}$ from $\partyset$ and reconstructs secret~$\MPCData{}{}$ using the sharing semantics of source~$\MPCSource$.
       \STATE {\bfseries share generation:} $\funcReshare$ computes $\MPCSharing{\MPCData{}{}}{\MPCTarget}$ from $\MPCData{}{}$ using sharing semantics of target~$\MPCTarget$.
        \STATE {\bfseries share distribution:} $\funcReshare$ sends the secret-shares of $\MPCSharing{\MPCData{}{}}{\MPCTarget}$ to parties in $\partyset'$ in accordance with the sharing semantics of target~$\MPCTarget$.
    \end{algorithmic}
    \end{small}
\end{algorithm}

The purpose of~$\funcReshare$ is to securely convert secret shares of data~$\MPCData{}{}$ from the sharing semantics of the source~$\MPCSource$, defined over a party set~$\partyset$~(i.e., $\MPCSharing{\MPCData{}{}}{\MPCSource}$) to those of the target~$\MPCTarget$, defined over a potentially different party set~$\partyset'$~(i.e., $\MPCSharing{\MPCData{}{}}{\MPCTarget}$).
The security guarantees of the~$\MPCReshare$ algorithm are inherently tied to the specific instantiation of~$\funcReshare$, which in turn depends on the threat models associated with both~$\MPCSource$ and~$\MPCTarget$.
In practice, these instantiations vary based on the deployment setting.
Detailed constructions and examples for clusters of~2, 3, and~4 parties are discussed in~\secref{sec:system-flow}.

\subsection{Threat Model}
\label{sec:framework-threat-model}
Given the hybrid nature of our framework, we adopt a~\emph{mixed} adversarial corruption model~\cite{AC:BJMS20,ITC:HirMul20} in~\FLname{}, assuming a centralized adversary~$\Adv$ that orchestrates the corruption\footnote{This is a stronger notion than having multiple independent adversaries.}. This setup allows the nature of corruption to vary across different MPC clusters, including the global server cluster. We formalize this via a \emph{corruption strategy}~$\AdvStrategy{\adventityset}$ defined over each entity set $\adventityset \in \{\globalserverset,\clusterserverset{i},\clientFLset{i}\}_{i \in \numclusters}$.

For instance, $\Adv$ may choose to maliciously corrupt certain MPC clusters~$\clusterserverset{i}$~(reflecting an unjustified trust assumption in the cluster), while corrupting others, including the global servers~$\globalserverset$, in a semi-honest manner. Within each entity set, $\Adv$ may corrupt a minority or a majority of parties, depending on the number of honest parties assumed (with the minimal assumption that each set contains at least one honest entity).

To reflect the hybrid structure of~\FLname{} composed of multiple~MPC clusters, we constrain the adversary~$\Adv$ to employ a~\emph{uniform corruption type} within each cluster -- either semi-honest or malicious -- but not a mixture of both. 
Allowing mixed corruption types within a single cluster~\cite{JC:Canetti00,AC:HirMauZik08}, while a natural extension, is left as future work.

\begin{table}[hb!]
    \centering
    \captionsetup{font=small}
    \caption{Possible configurations of~\FLnameTitle{} in terms of security against various corruption strategies of the adversary. Here $0 < t < \clusterserversize{}$.}
    \label{tab:my_label}
    \begin{small}
    \begin{tabular}{c r r r}
      \toprule
      Security   & Global: $\globalserverset$  & Clusters: $\clusterserverset{}$ & Clients: $\clientFLset{}$ \\
      \midrule
      I-A    & semi-honest    & $\clusterserversize{}$ semi-honest                   & semi-honest \\
      I-B    & semi-honest    & $t$ malicious       & semi-honest \\
      I-C    & semi-honest    & $\clusterserversize{}$ malicious                     & semi-honest \\
      \midrule
      II-A    & malicious     & $\clusterserversize{}$ semi-honest                   & semi-honest \\
      II-B    & malicious     & $t$ malicious       & semi-honest \\
      II-C    & malicious     & $\clusterserversize{}$ malicious                     & semi-honest \\
      \midrule
      III-A    & semi-honest    & $\clusterserversize{}$ semi-honest                   & malicious \\
      III-B    & semi-honest    & $t$ malicious       & malicious \\
      III-C    & semi-honest    & $\clusterserversize{}$ malicious                     & malicious \\
      \midrule
      IV-A    & malicious     & $\clusterserversize{}$ semi-honest                   & malicious \\
      IV-B    & malicious     & $t$ malicious       & malicious \\
      IV-C    & malicious     & $\clusterserversize{}$ malicious                     & malicious \\
      \bottomrule
    \end{tabular}
    \end{small}
\end{table}

The resulting security spectrum of~\FLname{} is summarized in~\tabref{tab:my_label}, ranging from the weaker semi-honest cases to the strongest all-malicious setting, depending on the adversary's corruption strategy across entities.

\myparagraph{$\funcFLname$ Ideal Functionality}
\label{sec:ideal-functionality}
We model the FL training process in~\FLname as an ideal functionality~$\funcFLname$ which is depicted in~\algref{alg:algorithm-ideal-functionality}.
For simplicity, we assume that the clients have secret-shared their training data with their respective~MPC clusters via the respective~$\MPCShare$ protocol.
In each iteration~$t$, $\funcFLname$ receives the shares of the current model~$\WWFLmodel{t-1}$ from the global servers~($\globalserverset$) and the shares of the training data from each cluster's servers in~$\clusterserverset{i}$ for~$i \in \numclusters$.
Using these shares, $\funcFLname$ reconstructs the current global model and the training data for each cluster.
Next, $\funcFLname$ trains the model~$\WWFLmodel{t-1}$ using the data from each cluster, resulting in a cluster model. 
These cluster models are then combined using an aggregation algorithm to obtain the updated global model~$\WWFLmodel{t}$ for this iteration.
Finally, $\funcFLname$ generates the secret-shares of~$\WWFLmodel{t}$ among the global servers~($\globalserverset$) and the cluster servers~($\clusterserverset{i}$) using their respective sharing semantics. 

\begin{algorithm}[hb!]
   \begin{small}
   \caption{$\funcFLname$: Ideal Functionality for~\FLnameTitle{}}
   \label{alg:algorithm-ideal-functionality}
    \begin{algorithmic}[1]
       \ACTORS $\funcFLname$, $\globalserverset, \clusterserverset{}$ 
       \COMMENT{$\clusterserverset{} = \bigcup_i \clusterserverset{i}$; $i \in \numclusters$}
       \INPUT $\MPCSharing{\WWFLmodel{t\mbox{-}1}}{}$, $\{\MPCSharing{\WWFLData{t}{\clusterserverset{i}}}{}\}_{\clusterserverset{i}\in\clusterserverset{}}$ \COMMENT{$\WWFLmodel{t\mbox{-}1}$ -- current global model, $\WWFLData{t}{\clusterserverset{i}}$ -- data at $\clusterserverset{i}$}
       \OUTPUT $\MPCSharing{\WWFLmodel{t}}{\globalserverset}$, $\MPCSharing{\WWFLmodel{t}}{\clusterserverset{i}}$\COMMENT{$\WWFLmodel{t}$ -- global model after iteration $t$} 
       \STATE {\bfseries initialize:} $\funcFLname$ receives $\MPCSharing{\WWFLmodel{t\mbox{-}1}}{}$ from $\globalserverset$ and reconstructs $\WWFLmodel{t\mbox{-}1}$.
       \FOR[in parallel]{each MPC cluster $\clusterserverset{i}\in\clusterserverset{}$}
            \STATE {\bfseries data reconstruction:} $\funcFLname$ receives $\MPCSharing{\WWFLData{t}{\clusterserverset{i}}}{}$ from $\clusterserverset{i}$ and reconstructs the cluster data $\WWFLData{t}{\clusterserverset{i}}$.\COMMENT{$\WWFLData{t}{\clusterserverset{i}}$ corresponds to collective data at $\clusterserverset{i}$}
            \STATE {\bfseries training:} $\funcFLname$ trains the current global model $\WWFLmodel{t\mbox{-}1}$ on $\WWFLData{t}{\clusterserverset{i}}$ to obtain the cluster model $\WWFLmodel{t}^{\clusterserverset{i}}$.
        \ENDFOR
        \STATE {\bfseries aggregation:} $\funcFLname$ aggregates $\{\WWFLmodel{t}^{\clusterserverset{i}}\}_{i \in \numclusters}$ using the respective secure aggregation algorithm to obtain $\WWFLmodel{t}{}$.
        \STATE {\bfseries secret-sharing (global servers):} $\funcFLname$ generates $\MPCSharing{\WWFLmodel{t}}{\globalserverset}$ from $\WWFLmodel{t}{}$.
        \STATE {\bfseries secret-sharing (cluster servers):} $\funcFLname$ generates $\MPCSharing{\WWFLmodel{t}}{\clusterserverset{i}}$ from~$\WWFLmodel{t}{}$ for $i \in \numclusters$.
    \end{algorithmic}
    \end{small}
\end{algorithm}

\myparagraph{Security of \FLnameTitle}
\label{sec:security-proof}
We adopt the standard stand-alone model of~\cite{JC:Canetti00}, assuming a static adversary and synchronous communication over perfectly secure channels\footnote{In practice, such communication channels can be instantiated via mutually authenticated~TLS.}.
Our security analysis is carried out in a simulation-based framework~\cite{BOOK:Lindell17}, which offers a modular and composable approach to proving security -- particularly well-suited to the clustered architecture of the~\FLname{} framework.

Specifically, for most subprotocols, we first define the corresponding~\emph{ideal functionalities}. These functionalities formalize their expected behavior in an ideal execution with a trusted third party, as is standard in~MPC.
Security of the overall framework is then established via a hybrid proof: each subprotocol securely realizes its functionality, and the composition theorem allows us to combine these results without re-deriving the security of every underlying instantiation~\cite{BOOK:Lindell17}.
Importantly, we do not assume the existence of corresponding~\enquote{ideal implementations}. Instead, we instantiate the functionalities using existing~MPC protocols with proven security guarantees, and compose them within~\FLname{} to implement more complex tasks.

Let~$\funcMPC^{\entityset}$ denote the set of~MPC ideal functionalities associated with the entity set~$\entityset \in \{\globalserverset,\, \clusterserverset{i}\}_{i \in \numclusters}$. These functionalities vary depending on the role of the entity in the \FLname{} framework. Specifically, for a standard FL setup,\footnote{In the presence of defense mechanisms, $\funcMPC^{\entityset}$ may include additional functionalities as listed in~\tabref{tab:mpc-functionalities}; these are straightforward generalizations.}
\[
\funcMPC^{\entityset} =
\begin{cases}
\{\funcShare^{\clusterserverset{i}},\, \funcReshare^{\clusterserverset{i}},\, \funcSample^{\clusterserverset{i}},\, \funcTrain^{\clusterserverset{i}}\} & \text{if } \entityset = \clusterserverset{i}, \\[0.5em]
\{\funcShare^{\globalserverset},\, \funcReshare^{\globalserverset},\, \funcAgg^{\globalserverset}\} & \text{if } \entityset = \globalserverset.
\end{cases}
\]
%
Given these definitions, the overall security of the~\FLname{} framework is formally stated in~Theorem~\ref{thm:funcFLnameSecurity}.

\begin{theorem}
    \label{thm:funcFLnameSecurity}
    Let~$\funcMPC^{\entityset}$ denote the set of ideal MPC functionalities for each entity set~$\entityset \in \{\globalserverset,\, \clusterserverset{i}\}_{i \in \numclusters}$. 
    Algorithm~$\FLTrain$~(\algref{alg:algorithm-workflow}) securely realizes the~$\funcFLname$ functionality~(\algref{alg:algorithm-ideal-functionality}) in the $\funcMPC^{\entityset}$-hybrid model, assuming a static adversary~$\Adv$ who corrupts the entities in~$\adventityset \in \{\globalserverset,\clusterserverset{i}, \clientFLset{i}\}_{i \in \numclusters}$, according to a predefined corruption strategy~$\AdvStrategy{\adventityset}$.
\end{theorem}

\begin{proofsketch}
    The~$\FLTrain$ algorithm~(\algref{alg:algorithm-workflow}) begins with the generation of secret shares of the initial global model~$\WWFLmodel{0}$ among the global servers~$\globalserverset$.
    This operation is performed by the designated stakeholder responsible for deploying the~\FLname{} framework, using the~$\MPCShare$ algorithm instantiated for~$\globalserverset$.
    The security of this sharing procedure is guaranteed by the ideal functionality~$\funcShare^{\globalserverset}$.

    In an iteration~$t \in [1,T]$, the global model~$\WWFLmodel{t-1}$, that is secret-shared among the global servers must be securely redistributed to the servers within each Layer~II cluster.
    This redistribution is carried out via the~$\MPCReshare$ protocol executed by~$\globalserverset$ with respect to each cluster~$\clusterserverset{i}$, for all~$i \in \numclusters$.
    The security of this step follows from the ideal functionality~$\funcReshare^{\globalserverset}$~(cf.~\algref{alg:algorithm-ideal-functionality-reshare}), where the source party set is~$\partyset = \globalserverset$, holding the secret-shared model~$\MPCSharing{\WWFLmodel{t-1}}{\globalserverset}$, and the target party set is~$\partyset' = \clusterserverset{i}$, which receives the reshared model~$\MPCSharing{\WWFLmodel{t-1}}{\clusterserverset{i}}$.

    In the next step of the~$\FLTrain$ algorithm, the selected clients~$\clientFL{i}{j}$ associated with cluster~$\clusterserverset{i}$~(i.e., those in the set~$\clientFLsetTotal{i}$ securely sampled via the~$\MPCSample$ algorithm) generate secret shares of their local data~$\WWFLData{t}{\clientFL{i}{j}}$ among the servers in~$\clusterserverset{i}$, denoted by~$\MPCSharing{\WWFLData{t}{\clientFL{i}{j}}}{\clusterserverset{i}}$.
    This sharing is performed using the~$\MPCShare$ algorithm instantiated for~$\clusterserverset{i}$, and its security is ensured by the corresponding ideal functionality~$\funcShare^{\clusterserverset{i}}$.
    Subsequently, each cluster~$\clusterserverset{i}$ performs~ML training on its collective dataset~$\MPCSharing{\WWFLData{t}{}}{\clusterserverset{i}}$ using the~$\MPCTrain$ algorithm, whose security is guaranteed by the corresponding ideal functionality~$\funcTrain^{\clusterserverset{i}}$.

    Even in the presence of malicious corruption, the ideal functionality~$\funcShare^{\clusterserverset{i}}$ guarantees that a malicious client cannot introduce inconsistent shares across the servers in~$\clusterserverset{i}$.
    While a malicious client may tamper with its input data prior to the secret-sharing process, this behavior falls outside the scope of~MPC itself and is typically addressed through complementary defense mechanisms~(cf.~\secref{sec:attacks_defenses} for defenses in~\FLname).
    As such, $\funcTrain^{\clusterserverset{i}}$ ensures the privacy and correctness of the ML training on data that has been secret-shared by clients.
    In cases where integrity violations arise due to adversarial behavior, the underlying secure aggregation protocol~$\MPCAgg$, which realizes the ideal functionality~$\funcAgg^{\globalserverset}$, plays a critical role in detecting and mitigating such attacks.

    In the next step, the local model~$\MPCSharing{\FLmodel{i}{t}}{\clusterserverset{i}}$ is securely reshared with the global servers~$\globalserverset$ using the~$\MPCReshare$ protocol.
    The security of this resharing step is ensured by the ideal functionality~$\funcReshare^{\clusterserverset{i}}$, where the source party set is~$\partyset = \clusterserverset{i}$, holding the secret-shared cluster model~$\MPCSharing{\FLmodel{i}{t}}{\clusterserverset{i}}$, and the target party set is~$\partyset' = \globalserverset$, which receives the reshared model~$\MPCSharing{\FLmodel{i}{t}}{\globalserverset}$.

    Finally, the global servers~$\globalserverset$ execute the~$\MPCAgg$ algorithm to securely aggregate the models received from all clusters, i.e., $\{\MPCSharing{\FLmodel{i}{t}}{\globalserverset}\}_{i \in \numclusters}$, and compute the updated global model in secret-shared form, denoted by~$\MPCSharing{\WWFLmodel{t}}{\globalserverset}$.
    The security of this aggregation step is guaranteed by the ideal functionality~$\funcAgg^{\globalserverset}$.

    Therefore, assuming that all underlying functionalities are securely instantiated in accordance with the specified security assumptions for each set~$\adventityset~\in~\{\globalserverset,\clusterserverset{i}, \clientFLset{i}\}_{i \in \numclusters}$ in~\FLname{}, the~$\FLTrain$ algorithm~(\algref{alg:algorithm-workflow}) securely realizes the ideal functionality~$\funcFLname$~(\algref{alg:algorithm-ideal-functionality}).
\end{proofsketch}

\subsection{Private Inference in \FLnameTitle}
\label{sec:framework-inference}
In~\FLname, after the defined number of training iterations~$T$ is completed, the~MPC clusters serve as clusters for~ML inference.
Here, we again utilize~PPML techniques to enable clients to privately query their clusters~\cite{NDSS:ChaudhariRS20,NeurIPS:KnottVHSIM21,EPRINT:MWCB22,NgC23,PETS:CSWYCA25,CHES:SanderBBE25}.
Consider the scenario where client~$\clientFL{}{}$ holding query~$\MPCQuery{}{}$ wants to use the inference service on a model~$\WWFLmodel{}$ that is secret-shared with a cluster~$\clusterserverset{k}$. 
This is accomplished by~$\clientFL{}{}$ generating~$\MPCSharing{\MPCQuery{}{}}{\clusterserverset{k}}$ using the~$\MPCShare$ algorithm instantiated for cluster~$\clusterserverset{k}$. Subsequently, the cluster servers in~$\clusterserverset{k}$ running~$\MPCInf$ on~$\MPCSharing{\WWFLmodel{}}{\clusterserverset{k}}$ and~$\MPCSharing{\MPCQuery{}{}}{\clusterserverset{k}}$ to generate the inference result in secret-shared form.
Finally, $\clusterserverset{k}$ reveals the result to~$\clientFL{}{}$ using the~$\MPCReveal$ protocol.

\subsection{Abstraction of Existing FL Schemes}
\label{sec:framework-abstraction}
The abstraction of \FLname{} captures multiple existing~FL frameworks, including standard~FL with a single aggregator~\cite{AISTATS:McMahanMRHA17}, distributed aggregators~\cite{TCHES:TXLZ25}, and hierarchical~FL schemes~\cite{IJCAI:Yang21}, as shown in \tabref{tab:abstraction}.
This abstraction not only simplifies comparisons but also facilitates advanced hybrid designs, including the integration of Differential Privacy~\cite{PETS:DP20,ACCESS:OA22}.

\begin{table}[htb!]
    \centering
    \captionsetup{font=small}
    \caption{Abstraction of existing~FL schemes~(cf.~\tabref{tab:relatedworksconcise}) in \FLnameTitle{}~(cf.~\figref{fig:architecture}). $\server$~--~aggregation server(s), S.Agg. -- secure aggregation, and~(S.)Agg. -- optional secure aggregation. See~\tabref{tab:notation} for other notations.}
    \label{tab:abstraction}
    \begin{small}
    \begin{tabular}{llcccc}
      \toprule
      \multicolumn{2}{l}{Scheme} & Layer I & Layer II & Layer III & Remark\\[0.4em]
      \midrule
      \multirow{2}[2]{*}{\makecell[l]{Aggregation\\(Single~$\server$)}}
      & $\sizeFL{s}$ & $\globalserversize=1$ & $\clusterserversize{i}=1$ & $\clusterclientsize{i}=1$ 
      & \multirow{2}{*}{$\clusterserverset{i}=\clientFLset{i}$}\\
      \cmidrule{2-5}
      & Role & (S.)Agg. & \multicolumn{2}{c}{ML Training} & \\[0.4em]
      \midrule
      \multirow{2}[2]{*}{\makecell[l]{Aggregation\\(Multi~$\server$)}}
      & $\sizeFL{s}$ & $\globalserversize>1$ & $\clusterserversize{i}=1$ & $\clusterclientsize{i}=1$ 
      & \multirow{2}{*}{$\clusterserverset{i}=\clientFLset{i}$}\\
      \cmidrule{2-5}
      & Role & S.Agg. & \multicolumn{2}{c}{ML Training} & \\[0.4em]
      \midrule
      \multirow{2}[2]{*}{\makecell[l]{Hierarchical\\FL}}
      & $\sizeFL{s}$ & $\globalserversize=1$ & $\clusterserversize{i}=1$ & $\clusterclientsize{i}>1$ 
      & \multirow{2}{*}{$\clusterserverset{i}\ne\clientFLset{i}$}\\
      \cmidrule{2-5}
      & Role & (S.)Agg. & (S.)Agg. & ML Training & \\[0.4em]
      \midrule
      \multirow{2}[2]{*}{\makecell[l]{\FLname\\\textbf{This Work}}}
      & $\sizeFL{s}$ & $\globalserversize>1$ & $\clusterserversize{i}>1$ & $\clusterclientsize{i}>1$ 
      & \multirow{2}{*}{$\clusterserverset{i}\ne\clientFLset{i}$}\\
      \cmidrule{2-5}
      & Role & S.Agg. & PPML Training & Data Sharing & \\
      \bottomrule
    \end{tabular}
    \end{small}
\end{table}

Standard~FL with a single aggregator~(\textsc{Single}~$\server$)~\cite{AISTATS:McMahanMRHA17} is a variant of~\FLname, where each~Layer~III cluster~$\clientFLset{i}$ consists of only one client that also serves as the~MPC cluster server~$\clusterserverset{i}$ in~Layer~II.
Thus, it is sufficient to conduct~ML training without global model privacy concerns and then send the results to a single global server~$\globalserverset$ in~Layer~I for aggregation.
The case of distributed aggregators~(\textsc{Multi}~$\server$)~\cite{TCHES:TXLZ25} follows similarly, except that secure aggregation is performed at Layer~I with multiple~($\globalserversize>1$) global servers.
Finally, existing hierarchical~FL schemes~\cite{IJCAI:Yang21} have a similar three-layer architecture as~\FLname, but use a single server at both the global and cluster level~($\globalserversize=1$, $\clusterserversize{i}=1$).
While~\FLname{} employs~PPML training at the cluster-server level, hierarchical~FL uses secure aggregation.
Additionally, clients in the hierarchical~FL approach perform local model training, as opposed to only data sharing in~\FLname.

Moreover, advanced hybrid designs in the~\FLname{} framework can accommodate \emph{privileged} clients---such as trusted consortium members---who may contribute data without requiring the global model to remain hidden. In such cases, the client can train the model locally and secret-share the resulting gradients directly with the global aggregation servers~$\globalserverset$.

\pagebreak

Another hybrid could involve collaborative training frameworks like SafeNet~\cite{SATML:CJO22}, which offer robustness against poisoning attacks. SafeNet constructs the global model as an ensemble of individually verified models, using majority voting to determine the final prediction. In the context of~\FLname{}, this setup can serve as a substitute for a conventional MPC cluster and its associated clients. The ensemble models can then be either aggregated using standard techniques or clustered via MPC-based methods~\cite{USENIX:NguyenRCYMFMMMZ22} to obtain a single representative model. From this point, the remaining operations conform to the~\FLname{} architecture, replacing SafeNet's plaintext operations with their MPC counterparts. Notably, training performed within these MPC clusters---which inherently resist poisoning---can be classified as \emph{attested training}. The global servers may then employ lightweight or mildly robust aggregation mechanisms to incorporate updates from these trusted clusters, thereby reducing the risk of error propagation to higher layers.

While our work primarily focuses on~MPC for~\FLname{} components, we emphasize that our abstraction is general enough to accommodate other privacy-preserving techniques, including those based on homomorphic encryption~(HE).
For example, as discussed in~\secref{sec:collaborative-learning-model-privacy}, existing multi-party~HE~(MHE) schemes preserve global model privacy but face scalability limitations~\cite{PETS:FTPSSB21,NDSS:SavPTFBSH21,TDSC:XuHXZLHD23}. 
Within~\FLname{}, such an~MHE-based setup can be treated as a single Layer~II cluster.
This allows the framework to leverage additional training data contributed by the~MHE setup, instead of excluding it entirely.
The main challenge in this integration lies in the~$\MPCReshare$ algorithm, which would require converting between~HE ciphertexts and~MPC-style secret shares.
Techniques proposed in works such as~MP2ML~\cite{ARES:BCD0Y20} and~Cheetah~\cite{USENIX:HLHD22} can potentially be adapted to support these conversions, enabling seamless interoperability within~\FLname{}.
We leave the integration of such~HE-based clusters into~\FLname{} as an interesting direction for future work.

\subsection{Example Configuration and Workflow in \FLnameTitle}
\label{sec:system-flow}
In this section, we outline the processing of the machine learning model within our \FLname framework~(cf.~\figref{fig:architecture}), using one specific configuration as an example from the various options listed in \tabref{tab:my_label}. This configuration, detailed in \tabref{tab:example_config}, involves three MPC clusters in Layer~II, each using a different MPC protocol and global servers using the 2PC + helper protocol in CrypTen~\cite{NeurIPS:KnottVHSIM21}. We use $\MPCSharing{x}{\entityset}$ to denote the secret shares of the value $x$ held by all the parties in the set $\entityset$. 

\begin{table*}[htb!]
    \centering
    \captionsetup{font=small}
    \caption{Example configuration for~\FLname with three~MPC Clusters in~Layer~II~(cf.~\figref{fig:architecture}). $\MPCSharing{x}{\entityset}$ represents the secret-shares of value~$x$ held by all the parties in the set~$\entityset$. In the~4PC setting~\cite{NDSS:KotiPRS22}, $\lv{x}{j}$ denotes the input-independent shares, and~$\mv{x}$ denotes the input-dependent share of~$x$.}
    \label{tab:example_config}
    \begin{small}
    \begin{tabular}{r r@{\hskip 0.25in} m{6.5cm}}
         \toprule
         MPC Servers & MPC Protocol & Details \\
         \midrule
         \makecell[r]{Layer~I\\Global Servers:~$\globalserverset$} & \makecell[r]{2PC + Helper\\\cite{NeurIPS:KnottVHSIM21}} & 
         \begin{tableitem}
             \item semi-honest security.
             \item $x = x_1 + x_2$.
             \item $\MPCSharing{x}{\globalserverset}: \left(\globalserver{1}: x_1, \globalserver{2}: x_2\right)$. 
         \end{tableitem}
         \vspace{-\baselineskip}\vfill
         \\
         \midrule
         \makecell[r]{Layer~II\\MPC Cluster 1:~$\clusterserverset{1}$} & 
         \makecell[r]{2PC Additive\\\cite{NDSS:Demmler0Z15}} & 
         \begin{tableitem}
             \item semi-honest security.
             \item $x = x_1 + x_2$.
             \item $\MPCSharing{x}{\clusterserverset{1}}: \left(\clusterserver{1}{1}: x_1, \clusterserver{1}{2}: x_2\right)$. 
         \end{tableitem}
         \vspace{-\baselineskip}\vfill
         \\
         \midrule
         \makecell[r]{Layer~II\\MPC Cluster 2:~$\clusterserverset{2}$} & 
         \makecell[r]{3PC Replicated\\\cite{CCS:AFLNO16,CCS:MohRin18}} & 
         \begin{tableitem}
             \item semi-honest security.
             \item $x = x_1 + x_2 + x_3$.
             \item
                \begin{tabular}[t]{@{}l@{\ }l@{}}
                \multirow{2}{*}{$\MPCSharing{x}{\clusterserverset{2}}$:} 
                    & $\clusterserver{2}{1}: (x_1, x_2)$, $\clusterserver{2}{2}: (x_2, x_3)$ \\
                    & $\clusterserver{2}{3}: (x_3, x_1)$ \\
                \end{tabular}      
         \end{tableitem}
         \\
         \midrule
         \makecell[r]{Layer~II\\MPC Cluster 3:~$\clusterserverset{3}$} & 
         \makecell[r]{4PC Replicated
         \\\cite{NDSS:KotiPRS22}} & 
         \begin{tableitem}
             \item malicious security.
             \item $x = \mv{x} - \lv{x}{1} - \lv{x}{2} - \lv{x}{3}$.
             \item
                \begin{tabular}[t]{@{}l@{\ }l@{}}
                \multirow{2}{*}{$\MPCSharing{x}{\clusterserverset{3}}$:} 
                & $\clusterserver{3}{1}: (\mv{x}, \lv{x}{1}, \lv{x}{2})$, $\clusterserver{3}{2}: (\mv{x}, \lv{x}{2}, \lv{x}{3})$,\\
                & $\clusterserver{3}{3}: (\mv{x}, \lv{x}{3}, \lv{x}{1})$, $\clusterserver{3}{4}: (\lv{x}{1}, \lv{x}{2}, \lv{x}{3})$.
                \end{tabular}
         \end{tableitem}
         \\
         \bottomrule
    \end{tabular}
    \end{small}
\end{table*}

The discussion follows the~\FLname training algorithm, detailed in~\algref{alg:algorithm-workflow}~(\FLTrain).
For simplicity, we omit the process of clients secret-sharing their training datasets with the Layer~II MPC clusters, as this can be done naively through the input-sharing protocol~$\MPCShare$~(see~\tabref{tab:mpc-functionalities}) of the underlying~MPC protocol.
Instead, we focus on the processing of the~ML model among the~MPC servers.

To enhance clarity in our discussion, we have divided the~\FLTrain{} algorithm into five steps, presented in the order of execution.
We describe the workflow throughout these steps, and for reference, we have included the corresponding line numbers from~\algref{alg:algorithm-workflow} in the format~${\sf L}123$.
For the remainder of this section, we assume that all operations are performed over a finite~$\ell$-bit ring, as required by the underlying~MPC protocols.
Additionally, to simplify the presentation, we omit the discussion of communication optimizations involving shared-key setups used in these protocols, and we refer the reader to~\cite{NeurIPS:KnottVHSIM21,NDSS:Demmler0Z15,CCS:AFLNO16,CCS:MohRin18,NDSS:KotiPRS22} for further details.

\subsubsection{Step I~(\alglineref{line:fltrain-step-a}): Initialization of Global Model}
In this step, the initial global model~$\WWFLmodel{0}$, likely owned by an organization or corporate entity, is secret-shared among the two global servers~$\globalserver{1}, \globalserver{2} \in \globalserverset$.
The model owner uses~CrypTen's~\cite{NeurIPS:KnottVHSIM21} input sharing protocol~$\MPCShare$ to secret-share~$\WWFLmodel{0}$, resulting in~$\MPCSharing{\WWFLmodel{0}}{\globalserverset}$, i.e., $\WWFLmodel{0} = \MPCSharing{\WWFLmodel{0}}{\globalserver{1}} + \MPCSharing{\WWFLmodel{0}}{\globalserver{2}}$.

\subsubsection{Step II~(\alglineref{line:fltrain-step-b}): Resharing of Global Model}
This step involves~\emph{resharing} the global model~$\WWFLmodel{t-1}$ for the current iteration $t$, to the three MPC clusters.
At a high level, each global server~$\globalserver{i}$ acts as an input party for the~MPC protocol within cluster~$\clusterserverset{j}$ and secret-shares its share of the global model.
Specifically, for global model~$x = \WWFLmodel{t-1}$, $\globalserver{i}$ secret-shares~$x_i = \MPCSharing{\WWFLmodel{t-1}}{\globalserver{i}}$, among the cluster servers~$\clusterserverset{j}$, represented as~$\MPCSharing{x_i}{\clusterserverset{j}}$. 

Due to the linearity of~MPC schemes, the cluster servers can locally add these shares.
This results in a secret sharing of~$x = \WWFLmodel{t-1}$ across the servers in cluster~$\clusterserverset{j}$, represented as~$\MPCSharing{x}{\clusterserverset{j}} = \MPCSharing{x_1}{\clusterserverset{j}} + \MPCSharing{x_2}{\clusterserverset{j}}$. 

\myparagraph{MPC Cluster~$\clusterserverset{1}$}
This cluster uses~2-out-of-2 additive secret sharing with semi-honest security~\cite{NDSS:Demmler0Z15}.
To generate~$\MPCSharing{x}{\clusterserverset{1}}$ with~$x = \WWFLmodel{t-1}$, the servers perform the following steps:
\begin{itemize}
    \item[--] Each~$\globalserver{i} \in \globalserverset$ samples random shares~$x_{i,1}$ and~$x_{i,2}$ such that~$x_i = x_{i,1} + x_{i,2}$. 
    \item[--] Each~$\globalserver{i}$ sends~$x_{i,j}$ to cluster server~$\clusterserver{1}{j} \in \clusterserverset{1}$.
    \item[--] Each~$\clusterserver{1}{j}$ locally computes~$\MPCSharing{x}{\clusterserver{1}{j}} = x_{1,j} + x_{2,j}$.
\end{itemize}

\myparagraph{MPC Cluster~$\clusterserverset{2}$}
This cluster is instantiated with a three-party replicated secret sharing with semi-honest security~\cite{CCS:AFLNO16,CCS:MohRin18}.
To generate~$\MPCSharing{x}{\clusterserverset{2}}$ with~$x = \WWFLmodel{t-1}$, the servers proceed as follows:
\begin{itemize}
    \item[--] Each~$\globalserver{i} \in \globalserverset$ samples random shares~$x_{i,1}$, $x_{i,2}$ and $x_{i,3}$ such that~$x_i = x_{i,1} + x_{i,2} + x_{i,3}$. 
    \item[--] $\globalserver{i}$ sends the pair $(x_{i,j},\, x_{i,j\%3+1})$ to $\clusterserver{2}{j} \in \clusterserverset{2}$ as its share in $\MPCSharing{x_i}{\clusterserverset{2}}$.
    \item[--] Servers in~$\clusterserverset{2}$ locally compute~$\MPCSharing{x}{\clusterserverset{2}} = \MPCSharing{x_1}{\clusterserverset{2}} + \MPCSharing{x_2}{\clusterserverset{2}}$. Specifically, each~$\clusterserver{2}{j}$ computes its share of~$\MPCSharing{x}{\clusterserverset{2}}$ as
    \[\MPCSharing{x}{\clusterserver{2}{j}} = (x_{1,j} + x_{2,j}, x_{1,j\%3+1} + x_{2,j\%3+1}).\]
\end{itemize}

\myparagraph{MPC Cluster~$\clusterserverset{3}$}
This cluster consists of four servers and utilizes the four-party~(4PC) replicated secret sharing with malicious security from~Tetrad~\cite{NDSS:KotiPRS22}.
Note that~Tetrad uses function-dependent preprocessing to achieve faster online phase. 
Consequently, the cluster servers in~$\clusterserverset{3}$ generate all the input-independent~$\lv{}{}$ values during the preprocessing phase.
Our discussion begins at this point, assuming that the preprocessing by the cluster servers has been completed.
From here, the servers proceed as follows:
\begin{itemize}
    \item[--] Cluster servers send the following~$\lv{}{}$ shares to~$\globalserver{i} \in \globalserverset$:
    \begin{align*}
        \clusterserver{3}{1}, \clusterserver{3}{3}, \clusterserver{3}{4} &: \lv{x_i}{1}\\
        \clusterserver{3}{1}, \clusterserver{3}{2}, \clusterserver{3}{4} &: \lv{x_i}{2}\\
        \clusterserver{3}{2}, \clusterserver{3}{3}, \clusterserver{3}{4} &: \lv{x_i}{3}
    \end{align*}
    \item[--] $\globalserver{i}$ accepts the~$\lv{}{}$ values that form the majority and computes~$\lv{x_i}{} = \lv{x_i}{1} + \lv{x_i}{2} + \lv{x_i}{3}$.
    \item[--] $\globalserver{i}$ sends~$\mv{x_i} = x_i - \lv{x_i}{}$ to~$\clusterserver{3}{1}, \clusterserver{3}{2}$, and~$\clusterserver{3}{3}$, thus completing the input sharing to generate~$\MPCSharing{x_i}{\clusterserverset{3}}$. 
    \item[--] Servers in~$\clusterserverset{3}$ locally compute~$\MPCSharing{x}{\clusterserverset{3}} = \MPCSharing{x_1}{\clusterserverset{3}} + \MPCSharing{x_2}{\clusterserverset{3}}$.
\end{itemize}

\subsubsection{Step III~(\alglineref{line:fltrain-step-c}): Cluster Training of Local Model}
This step involves~MPC-based~ML training at the cluster servers, using secret shares of the current global model, represented as~$\MPCSharing{\WWFLmodel{t-1}}{\clusterserverset{j}}$, along with the data from the clients associated with the cluster in the current iteration, denoted by~$\MPCSharing{\WWFLData{t}{}}{\clusterserverset{i}}$.
The servers then execute the underlying~PPML training algorithm, $\MPCTrain$, and obtain secret shares of the updated model, denoted by~$\MPCSharing{\WWFLmodel{t}}{\clusterserverset{j}}$.

\subsubsection{Step IV~(\alglineref{line:fltrain-step-d}): Resharing of Local Model}
Once each cluster receives its updated local model for iteration~$t$ in secret-shared form~$\MPCSharing{\WWFLmodel{t}}{\clusterserverset{j}}$, the next step is to reshare this model among the global servers~$\globalserverset$.
This process is similar to~Step~II above, with the key difference being that the roles of the source and destination of the secret sharing are now reversed. 
We detail the steps for each of the three~Layer~II clusters next.

\myparagraph{MPC Cluster~$\clusterserverset{1}$}
To generate~$\MPCSharing{x}{\globalserverset}$ with~$x = \FLmodel{1}{t}$, the cluster servers in~$\clusterserverset{1}$ perform the following steps:
\begin{itemize}
    \item[--] Each~$\clusterserver{1}{i} \in \clusterserverset{1}$ samples random shares~$x_{i,1}$ and~$x_{i,2}$ such that~$x_i = x_{i,1} + x_{i,2}$. 
    \item[--] Each~$\clusterserver{1}{i}$ sends~$x_{i,j}$ to global server~$\globalserver{j} \in \globalserverset$.
    \item[--] Each~$\globalserver{j}$ locally computes~$\MPCSharing{x}{\globalserver{j}} = x_{1,j} + x_{2,j}$.
\end{itemize}

\myparagraph{MPC Cluster~$\clusterserverset{2}$}
The resharing process in cluster~$\clusterserverset{2}$, which uses a three-party replicated sharing scheme where~$x = \FLmodel{2}{t}$ is represented as~$x = x_1 + x_2 + x_3$, closely resembles that of cluster~$\clusterserverset{1}$ with a few minor differences.

First, cluster~$\clusterserverset{2}$ involves three servers, $\clusterserver{2}{i}$, compared to the two servers in~$\clusterserverset{1}$.
Consequently, each global server~$\globalserver{j}$ will receive secret shares of three values instead of two.
Second, the local computation for a secret-shared value~$x$ in~$\clusterserverset{2}$ involves adding three shares: $\MPCSharing{x}{\globalserver{j}} = x_{1,j} + x_{2,j} + x_{3,j}$, rather than just two as in~$\clusterserverset{1}$.

Additionally, because~$\clusterserverset{2}$ operates under semi-honest security, an optimization can be applied. Specifically, $\clusterserver{2}{1}$ can directly secret-share the sum~$(x_1 + x_2)$, while~$\clusterserver{2}{2}$ secret-shares~$x_3$, effectively eliminating one instance of secret-sharing.

\myparagraph{MPC Cluster~$\clusterserverset{3}$}
Since this cluster uses a maliciously secure four-party protocol from~Tetrad~\cite{NDSS:KotiPRS22}, we cannot directly follow the approach where each cluster server~$\clusterserver{3}{i}$ secret-shares its share of~$x = \FLmodel{3}{t}$ with the global servers in~$\globalserverset$.
However, because the protocol uses replicated secret sharing, each share is held by three of the four parties, with at most one being corrupt.
Therefore, instead of a single cluster server sharing its own secret, the share can be~\enquote{jointly} shared by three cluster servers.
This ensures that a majority of correct values is always available at the global server~$\globalserver{j}$.
The remaining steps are similar to those for other clusters and are described below.

\begin{itemize}
    \item[--] Cluster servers~\enquote{jointly} secret share the following shares among the two global servers in~$\globalserverset$:
    \begin{align*}
        \clusterserver{3}{1}, \clusterserver{3}{3}, \clusterserver{3}{4} &: \lv{x}{1}\\
        \clusterserver{3}{1}, \clusterserver{3}{2}, \clusterserver{3}{4} &: \lv{x}{2}\\
        \clusterserver{3}{2}, \clusterserver{3}{3}, \clusterserver{3}{4} &: \lv{x}{3}\\
        \clusterserver{3}{1}, \clusterserver{3}{2}, \clusterserver{3}{3} &: \mv{x}
    \end{align*}
    \item[--] For each of the four instances, $\globalserver{i}$ accepts the share that forms the majority.
    \item[--] Servers in~$\globalserverset$ locally compute~$\MPCSharing{x}{\globalserverset} = \MPCSharing{\mv{x}}{\globalserverset} - \MPCSharing{\lv{x}{1}}{\globalserverset} - \MPCSharing{\lv{x}{2}}{\globalserverset} - \MPCSharing{\lv{x}{3}}{\globalserverset}$.
\end{itemize}

\subsubsection{Step V~(\alglineref{line:fltrain-step-e}): Secure Aggregation}
In this step, the global servers perform secure aggregation~$\MPCAgg$ using the cluster models that are secret-shared among them.
In our example, these secret shares are$\{\MPCSharing{\FLmodel{1}{t}}{\globalserverset}, \MPCSharing{\FLmodel{2}{t}}{\globalserverset}, \MPCSharing{\FLmodel{3}{t}}{\globalserverset}\}$, as obtained in Step~IV.
After the~$\MPCAgg$ protocol, the global servers~$\globalserverset$ receive secret shares of the updated global model for iteration~$t$, denoted by~$\MPCSharing{\WWFLmodel{t}}{\globalserverset}$. 
This updated model will serve as the starting point for iteration~$t+1$.

In this example, the global servers are implemented using~CrypTen~\cite{NeurIPS:KnottVHSIM21}, and the secure aggregation can be executed using the~\FLname{} code, which provides an implementation of secure aggregation in~CrypTen.
However, the global servers are not limited to~CrypTen.
They can also be instantiated with other~MPC protocols.
For example, \textsc{SafeFL}~\cite{DLS:GMSSWY23} offers a~PyTorch connector plugin for using~PyTorch with the~MP-SPDZ~\cite{CCS:Keller20} framework, which supports a range of efficient~MPC protocols.

Similarly, while we use three distinct~MPC protocols with a small number of parties to instantiate the MPC clusters in~Layer~II, our framework is not limited to these specific protocols. 
The technique of resharing secret shares is well-known in the~MPC literature and has been used in several works such as~\cite{CHES:AarajAGMMPSSSSS25,ARES:BCD0Y20}, though it is less common in the context of federated learning.
For example, the~\enquote{converter unit} in~FANNG-MPC~\cite{CHES:AarajAGMMPSSSSS25} introduces methods to convert secret shares from one set of~$n$ parties in a dishonest majority setting with malicious security to another set of~$m \ne n$ parties in the same corruption setting.
Similarly, MP2ML~\cite{ARES:BCD0Y20} presents a scheme for converting between~HE~(using the~CKKS scheme) and~MPC to enable private~ML inference.
These works highlight the flexibility of our~\FLname{} framework, which enables global federated learning and integrates various existing techniques. 
\section{Performance Evaluation}
\label{sec:evaluation}
We systematically and empirically evaluate the performance of \FLname in terms of accuracy as well as computation and communication overhead.
We use standard image classification tasks and build a prototype implementation\footnote{\url{https://encrypto.de/code/WW-FL}\label{refnote}} using the~MPC framework CrypTen~\cite{NeurIPS:KnottVHSIM21}.
Future research can extend our evaluation to stronger security models (cf.~\S\ref{sec:framework-threat-model}), as well as more sophisticated ML models and training tasks.

\subsection{Implementation}
\label{sec:prototype_implementation}
We implement~\FLname based on the~CrypTen MPC framework developed by Meta~\cite{NeurIPS:KnottVHSIM21}.
CrypTen provides a PyTorch-style interface and implements~MPC operations with~GPU support.
Specifically, it implements semi-honest arithmetic and~Boolean secure two- and multi-party computation protocols that use a trusted third~\enquote{helper} party to generate correlated randomness.
CrypTen provides a~\enquote{simulation} mode where the specified computation is performed on a single node in plaintext yet simulates all effects that computation in~MPC would have on the results~(e.g., due to limited fixed-point precision and truncation).
We leverage this mode to efficiently evaluate~\FLname's accuracy and later the impact of data-poisoning attacks. For our run-time and communication measurements, we benchmark the full MPC protocols.
In all our experiments, the fixed-point precision in~CrypTen is set to~22 decimal bits~(the maximum value recommended by the developers).

We use~CrypTen to implement~(a) private training on~Layer~II and~(b)~distributed aggregation on~Layer~I.
CrypTen out of the box supports private inference between~Layer~II and~III.
We extend~CrypTen with an identity layer to enable model conversions and re-sharing.
Additionally, we extend the implementation of convolutional layers to enable full~GPU-accelerated training for such model architectures.
Moreover, we provide the necessary code to orchestrate the various parties and components, thereby creating a unified simulation framework. 

Moving our prototype implementation to production requires hardening both the cryptographic and system layers.
For example, all processing of sensitive data~(even when secret-shared) must happen in constant time to protect against timing-based side-channel attacks.
Network connections must be secured, e.g., via mutually authenticated~TLS to instantiate communication channels that are assumed to be perfectly secure.
Finally, servers must be hardened, their~APIs must be secured, and reliable key and identity management must be implemented.
Please refer to recommended practice for~MPC-based systems defined by~ITU-T~X.1770~\cite{ITU} and~IEEE~P2842~\cite{IEEE} for details.

\myparagraph{Setup}
\label{sec:prototype_setup}
Plaintext~FL and~CrypTen-based~\FLname~\emph{simulations} are run on a single computing platform with two~Intel Xeon Platinum~8168~CPUs, 1.5TB~RAM, and~16~NVIDIA Tesla~V100 GPUs.
For realistic results in a~\emph{distributed~MPC deployment} with two computational and one helper party, we use three~Amazon~AWS~g3s.xlarge instances with~4 vCPUs, and~8GB~GPU memory on an~NVIDIA Tesla~M60.
These instances are located in the same~AWS availability zone~(due to high costs associated with routing traffic between different zones), yet we simulate intra- and inter-continental network connections by setting the bandwidth to~1Gbps/100Mbps and latency to~20ms/100ms respectively.

\myparagraph{Tasks}
\label{sec:prototype_tasks}
Following prior work such as~\cite{SP:RatheeSWP23,SATML:BMPSTVWYY24,TCHES:TXLZ25}, we evaluate~\FLname on two standard image classification tasks: recognizing~(a)~hand-written digits using LeNet~\cite{CHES:AarajAGMMPSSSSS25} trained on~MNIST~\cite{DBLP:journals/pieee/LeCunBBH98} dataset, and~(b)~objects in one of~10 classes using~ResNet9 trained on~CIFAR10~\cite{Krizhevsky_2009_17719} dataset.
We simulate~1000 clients overall from which~100 are randomly sampled during each training iteration. In the~FL setting, all sampled clients provide their computed update directly to the aggregation server.
In the~\FLname setting, each of the~10~MPC clusters has~100 associated clients from which each samples~10 clients at random. Each client has~200 data points assigned to it at random with duplicates allowed between clients.

For plain~FL, we use batch size~8 and learning rate~0.005, and train locally for~5 epochs before central aggregation.
To allow for a fair comparison between~FL and~\FLname, we use a correspondingly~\emph{scaled} batch size of~80 and a learning rate of~0.05 as in~\cite{ARXIV:GDGNWKTJH17}. 
Each client and~MPC cluster performs~5 local training epochs with either a batch size of~8 or~80, respectively, following the recommendation in~\cite{ARXIV:GDGNWKTJH17}.
Tab.~\ref{tab:basic_run_config} summarises all the~(hyper) parameters used during the training.

\begin{table}[htb!]
\centering
\captionsetup{font=small}
\caption{Training parameters used in~FL and~\FLnameTitle.}
\label{tab:basic_run_config}
\begin{small}
\begin{tabular}{llr}
\toprule
Parameter & FL & \FLname \\
\midrule
    \# clients & \multicolumn{2}{c}{1000}\\
    \# clients selected per round & \multicolumn{2}{c}{100}\\
    \# MPC clusters & - & 10\\
    \# clients per MPC cluster & - & 100\\
    \# clients per MPC cluster per round & - & 10\\
    size of client datasets & \multicolumn{2}{c}{200}\\
    learn rate & 0.005 & 0.05\\
    \# local epochs &\multicolumn{2}{c}{5}\\
    batch size & 8 & 80\\
\bottomrule
\end{tabular}
\end{small}
\end{table}

\subsection{Results}
In this section, we answer the following research questions to empirically evaluate performance and discuss our results:
\begin{myitem}%
    \item[Q1:] What is the accuracy difference between~FL~\cite{AISTATS:McMahanMRHA17} and~\FLname (in plaintext)?
    \item [Q2:] What is the impact on accuracy for~\FLname when moving from plaintext to MPC?
    \item [Q3:] What are the run-time and communication overheads of (MPC-based) \FLname compared to~FL?
    \item [Q4:] How does~\FLname scale and compare to existing fully private training approaches?
\end{myitem}

\begin{figure}[htb!]
    \centering
    \subfigure[LeNet trained on MNIST for 500 epochs.]{
        \includegraphics[width=0.45\textwidth]{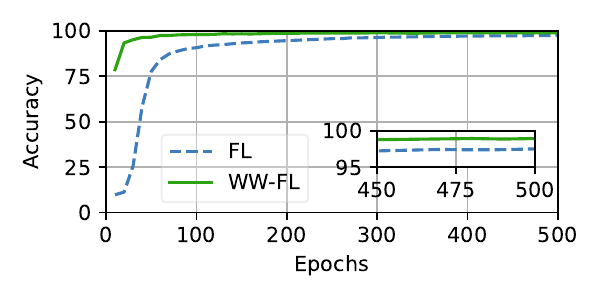}}
    \subfigure[ResNet9 on CIFAR10 for 500 epochs.\label{fig:FL-HyFL-FedAvg-500-no-attack}]{\includegraphics[width=0.45\textwidth]{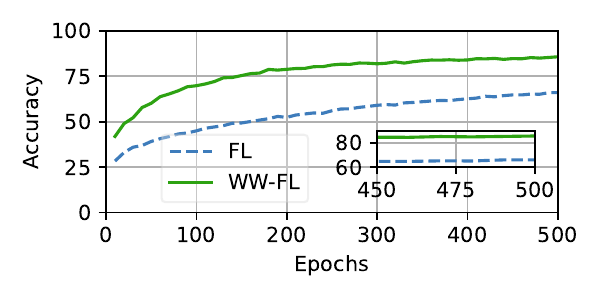}
    }
    
    \vspace{-2mm} 
    
    \subfigure[LeNet trained on MNIST for 2000 epochs.]{
        \includegraphics[width=0.45\textwidth]{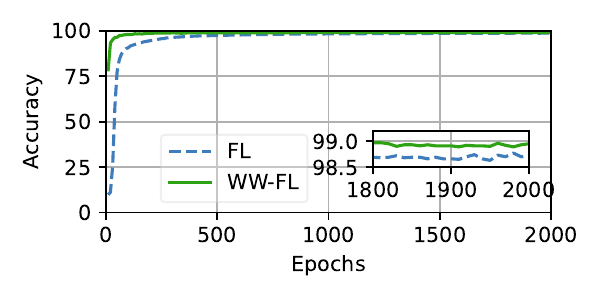}}
    \subfigure[ResNet9 on CIFAR10 for 2000 epochs.\label{fig:FL-HyFL-FedAvg-2000-no-attack}]{\includegraphics[width=0.45\textwidth]{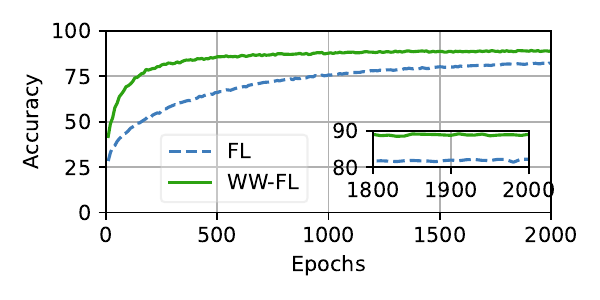}
    }
    \vspace{-2mm}
    \captionsetup{font=small}
    \caption{Validation accuracy for FL and \FLnameTitle{}.}
    \label{fig:FL-HyFL-FedAvg}
\end{figure}

\myparagraph{Q1 -- FL vs~\FLname}
In~Fig.~\ref{fig:FL-HyFL-FedAvg}, we compare the validation accuracy of~FL and~\FLname for image classification tasks for~500 and 2000~rounds.
Here, \FLname converges significantly faster than regular~FL, e.g., after~500 rounds of training~ResNet9 on~CIFAR10, \FLname reaches 85.68\% validation accuracy, whereas regular~FL only reaches~65.95\%. Similarly, for~MNIST, \FLname achieves~98.95\% accuracy compared to~98.72\% for regular~FL.
We attribute this to~\FLname pooling training data at the cluster level and thus being able to exploit the known benefits of batching~\cite{ARXIV:GDGNWKTJH17,SIAM:BCN18,AAAI:WangWCJ22}.

\begin{figure}[htb!]
    \centering
    \includegraphics[width=0.49\textwidth]{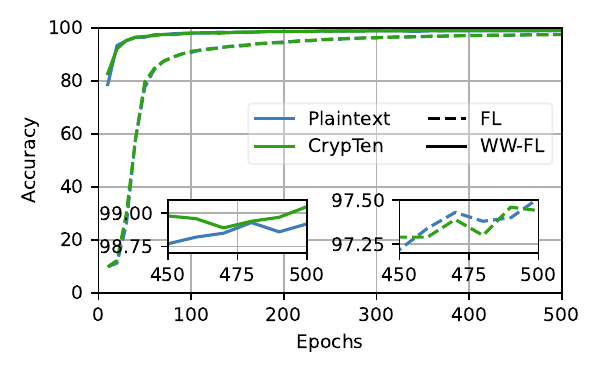}
    \vspace{-3mm}
    \captionsetup{font=small}
    \caption{Validation accuracy for~FL and~\FLnameTitle{} in plaintext and~MPC~(CrypTen simulation) for~LeNet/MNIST training.}%
    \label{fig:FL-HyFL-FedAvg-500-2000-no-attack-Crypten-Comparison}
    \vspace{-2mm}
\end{figure}

\myparagraph{Q2 -- Accuracy Impact of~MPC}
In~Fig.~\ref{fig:FL-HyFL-FedAvg-500-2000-no-attack-Crypten-Comparison}, we compare the plaintext validation accuracy~(cf.~Q1) to our~CrypTen simulation to measure the impact of~MPC~(i.e., fixed-point representation with~22 bit decimal representation and truncation). While there is a slight difference in initial rounds, both implementations quickly converge to almost the same validation accuracy, with only a small difference on the order of~0.1\%.

\myparagraph{Q3 -- MPC Overhead}
Finally, we evaluate the overhead of~MPC for secure training and aggregation.
For this, we measure the run-times and communication for one iteration of~LeNet/MNIST training~(i.e., 5 local epochs) in~AWS for one cluster~(with~1Gbps bandwidth and~20ms latency) and one iteration of global aggregation~(with 100Mbps bandwidth and~100ms latency).
The training for~\FLname on the cluster level takes~315s and requires~5.28GB inter-server communication, which is multiple orders of magnitude more than local plaintext training in~PyTorch~(which only takes~0.07s).
The aggregation over~10 cluster inputs is very efficient with~0.023s run-time and has no communication overhead since only linear operations are required, which are local operations in~MPC.

Additional overhead that must be considered for clients is sharing data with the training cluster servers.
In our setup, clients on expectation have to upload~3.31MB and~9.86MB in total for~500 rounds of training for~MNIST and~CIFAR10, respectively.
Furthermore, we have to account for sharing the trained models from training clusters to the aggregation servers.
Given the number of model parameters and~CrypTen sharing semantics, each training cluster must transfer~0.49MB and~39.19MB per server for~LeNet and~ResNet9, respectively.
This clearly shows that it is significantly more efficient for participants to upload their training data in secret-shared form compared to down- and uploading model parameters for each training round.
In our evaluation setup, the training clusters and the aggregation layer use the same~MPC configuration, hence no interactive re-sharing is necessary.

\myparagraph{Q4 -- Scalability \& Comparison}
The only existing solutions for privacy-preserving training and inference with cryptographic guarantees and global model privacy are based on either~MPC or~MHE~(cf.~\secref{sec:prelims}).
Existing~MPC-based~PPML solutions with global model privacy can be seen as a special case of~\FLname with a single~MPC cluster on~Layer~II and no aggregation at~Layer~I~(cf.~\tabref{tab:abstraction}).
In terms of~MHE, \Spindle~\cite{PETS:FTPSSB21}, \Poseidon~\cite{NDSS:SavPTFBSH21}, and~\Hercules~\cite{TDSC:XuHXZLHD23} are currently the most promising solutions for collaborative training and inference with global model privacy.
We therefore compare scalability and especially training overheads with these existing approaches.

For the comparison with~MPC, we first assume there is a fixed threshold on the communication overhead a single~MPC training cluster can handle in a single round.
For example, we can assume a maximum of~$\approx$5GB of inter-server communication.
Based on our experimental results provided for~Q3, the maximum number of clients a centralized~MPC-based~PPML system can handle is~$\approx$10.
In contrast, in~\FLname with our example configuration of~10 independent~MPC training clusters and one global aggregation cluster, we can easily handle~10$\times$ more clients~(100 randomly selected per round).
Here, the number of supported clients scales linearly in the number of available independent~MPC training clusters.
\emph{As such, moving from a single~MPC training cluster to our~\FLname architecture is warranted whenever the resources of a single cluster are exceeded; this can be as low as~10 clients in our example}.
When considering computation overhead, the run-time for a single~MPC training cluster~(and therefore for a centralized~MPC-based~PPML system) scales linearly in the number of clients and training samples.
In contrast, in~\FLname, through adding new~MPC training clusters, additional workload resulting from newly joining clients or an increasing number of training samples can be trivially parallelized, thereby not significantly increasing total training run-time; the only minor increase results from the linearly scaling global aggregation step, which however is at least four orders of magnitude more efficient than training -- even with full privacy protection~(cf.~run-time measurements provided for~Q3). 

For the comparison with~MHE, we refer to the experiments reported in~\cite{PETS:FTPSSB21,NDSS:SavPTFBSH21,TDSC:XuHXZLHD23}.
\Spindle's~\cite{PETS:FTPSSB21} communication overhead is demonstrated to scale linearly in the number of data providers~(Layer~III clients in our terminology) and ciphertext size, which in turn depends on the number and dimension of training samples.
This is similar to~\FLname, however, we can easily handle additional load from new clients by adding~MPC training clusters, thus not increasing the load for existing clusters or clients.
Furthermore, \Spindle only supports linear and logistic regression models, whereas we validate~\FLname with more complex convolutional neural networks.

\Poseidon~\cite{NDSS:SavPTFBSH21} extends and improves upon~\Spindle: while it also scales linearly in the number of clients and training samples, it supports various neural network architectures.
However, it relies on approximations for non-linear layers due to the restrictions of~HE schemes.
Specifically, the evaluation includes a~3-layer fully-connected neural network for~MNIST classification that achieves~89.9\% accuracy after~1000 epochs, whereas we validate~\FLname with a network that includes three convolutional and two fully-connected layers~(along with non-linear activation functions and pooling layers)~\cite{CHES:AarajAGMMPSSSSS25}, and surpasses~90\% accuracy in less than~100 rounds with~5 local epochs in each round~(cf.~results reported for~Q1).
More comparable is the evaluation of~\cite{NDSS:SavPTFBSH21} on a model for~CIFAR10 classification with two convolution and two fully-connected layers, which we therefore reference in our comparison in~\tabref{tab:scalability}.
Another notable benchmark of~\Poseidon~\cite{NDSS:SavPTFBSH21} is training the three-layer~SecureML neural network~\cite{SP:MohasselZ17} with three clients for~15 epochs on~MNIST, which takes~73.1 hours, whereas~MPC frameworks such as~\Falcon~\cite{PoPETS:WTBKMR21} only require~0.56 hours for the same task.

\Hercules~\cite{TDSC:XuHXZLHD23} advances the state of the art over~\Poseidon by enabling more efficient homomorphic matrix operations together with polynomial approximations for non-linear activation functions. 
It further demonstrates the feasibility of training with up to 50 clients, achieving secure aggregation without relying on multiple non-colluding servers and providing strong privacy guarantees under the passive security model. 
These results highlight the promise of multi-party homomorphic encryption~(MHE)-based approaches for privacy-preserving federated learning. 
Nevertheless, MHE-based solutions also bring certain costs: they require a multi-party key setup among all clients, entail sizable ciphertexts that increase storage requirements, and typically only support passive security while lacking dedicated mechanisms against poisoning attacks. 
For instance, training with 10 clients in~\Hercules on~MNIST requires a total storage of~49.06GB across the clients. 
By contrast, \FLname adopts an orthogonal design point: clients only store their training data in plaintext, while servers hold additive shares of training data and model parameters, resulting in less than~4GB of storage in total. 
On the downside, however, \FLname using naive MPC alone requires the availability of at least two non-colluding servers in every MPC cluster. 
Moreover, while private training is efficiently parallelized across clusters, the overhead compared to cleartext federated learning becomes significant when considering the overall cost of achieving global model privacy.

In~\tabref{tab:scalability}, we summarize performance results for the approaches compared above.
For the total run-time and communication overhead for~\FLname, we count data sharing between~Layer~III clients and~Layer~II cluster servers, Layer~II training, secret-shared parameter up- and download between~Layer~II and Layer~I cluster servers, and secure aggregation on~Layer~I.
For the centralized~MPC-based method, we only have to consider data sharing between clients and one~MPC cluster with the same configuration as a~\FLname~Layer~II cluster that executes all training epochs on its own.
While this results in slightly reduced communication overhead compared to~\FLname~(as secure aggregation can be omitted), there is an order of magnitude increase in run-time as the training procedure cannot be parallelized as for~\FLname.
Also, note that it might not be practically feasible for a single cluster to handle the substantial total communication overhead that is evenly distributed among all clusters in~\FLname.

Finally, we report the original benchmark results of~\Poseidon~\cite{NDSS:SavPTFBSH21} and~\Hercules~\cite{TDSC:XuHXZLHD23}.
Due to the severely limited scale of experiments in terms of supported number of clients as well as the limited model architecture complexity, a fair comparison to~\FLname is not entirely feasible.
However, when relying on the theoretical linear scalability in terms of client numbers and training epochs, we can estimate that~\Hercules would require roughly~1.86$\times$ more run-time overhead than~\FLname for their~\enquote{CIFAR-10-N1} model that is somewhat comparable yet still less complex than~LeNet.

\begin{table}[htb!]
\centering
\captionsetup{font=small}
\caption{Comparison of total run-time in hours, communication in terabytes, and storage in gigabytes for~\FLname and central~MPC-based training for~1000 clients, and~MHE-based approaches~(\Poseidon~\cite{NDSS:SavPTFBSH21} and~\Hercules~\cite{TDSC:XuHXZLHD23}) for~10 and~50 clients. Note that the neural networks trained for~\Poseidon and~\Hercules are less complex and do not include non-linear layers. Type of layers: CV -- convolution, MP -- max pooling, FC -- fully connected, AP -- average pooling.}
\label{tab:scalability}
\resizebox{\linewidth}{!}{
{\begin{tabular}{|l|r|r|r|r|r|r|r|}
\hline
Metric & \FLname & MPC~\cite{NeurIPS:KnottVHSIM21} & \cite{NDSS:SavPTFBSH21} & \cite{TDSC:XuHXZLHD23} & \cite{NDSS:SavPTFBSH21} & \cite{TDSC:XuHXZLHD23}\\
\hline
    \# clients  & \numprint{1000} & \numprint{1000} & 10 & 10 & 50 & 50 \\
    \# clients per round  & 100 & 100 & 10 & 10 & 50 & 50 \\
    \# epochs in total  & \numprint{2500} & \numprint{2500} & \numprint{1000} & \numprint{1000} & \numprint{25000} & \numprint{25000} \\
    Accuracy  & 98.95\% & 98.95\% & 88.7\% & 91.8\% & 51.8\% & 54.3\% \\\hline
    \multirow{2}{*}{Model} & \multicolumn{2}{c|}{LeNet~\cite{CHES:AarajAGMMPSSSSS25}:} & \multicolumn{2}{c|}{\enquote{MNIST}~\cite{TDSC:XuHXZLHD23}:} & \multicolumn{2}{c|}{\enquote{CIFAR-10-N1}~\cite{TDSC:XuHXZLHD23}:} \\
     & \multicolumn{2}{c|}{3CV, 4ReLU, 2MP, 2FC} & \multicolumn{2}{c|}{3FC} & \multicolumn{2}{c|}{2CV, 1AP, 1MP, 2FC} \\
    \hline
    Total run-time~(in h)  & \numprint{43.83} & \numprint{437.57} & \numprint{1.24} & \numprint{0.43} & \numprint{126.26} & \numprint{40.73} \\
    Total comm.~(in TB) & \numprint{26.45} & \numprint{26.40} & \numprint{7.03} & \numprint{0.18} & \numprint{3076.17} & \numprint{102.54} \\
    Total storage~(in GB) & \numprint{3.39} & \numprint{3.33} & \numprint{3140.25} & \numprint{49.06} & \numprint{105468.5} & \numprint{823.50} \\
\hline
\end{tabular}}
}
\end{table}
\section{Attacks \& Mitigations}
\label{sec:attacks_defenses}
In FL, malicious participants can degrade accuracy through data or model poisoning attacks~\cite{ACMCS:ZLJS22}. Because \FLname models are not accessible to clients, these attackers are limited to manipulating the training data they provide, resulting in either targeted, backdoor, or untargeted attacks. Our work is focused on untargeted~FL poisoning, as it is most relevant to real-world applications and more difficult to detect~\cite{USENIX:FangCJG20,SP:SHKR22,SP:RatheeSWP23,TCHES:TXLZ25}.
We therefore propose how to systematically evaluate the effectiveness of state-of-the-art data-poisoning attacks~\cite{ECAI:XiaoXE12,USENIX:FangCJG20,ESORICS:TolpeginTGL20,SP:SHKR22} in the~\FLname setting 
and propose a new robust aggregation scheme as a possible mitigation.

\subsection{Data-Poisoning Attacks in \FLnameTitle}
\label{sec:data-poisoning-attacks}
In data-poisoning attacks, malicious clients can perform arbitrary manipulations on the training data. State-of-the-art attacks are based on label flipping, where clients keep the legitimate training samples, yet exchange the associated labels according to different strategies~\cite{USENIX:FangCJG20, ESORICS:TolpeginTGL20,SP:SHKR22}.

To evaluate~\FLname{}, we implemented four label-flipping attacks detailed below. We assume a single attacker controlling all malicious clients, coordinating their actions. Only one attack occurs at a time, and if multiple clients have the same data sample, their poisoned labels will be the same. The assumption is that each malicious client poisons all its data samples to maximize the attack's impact. The data is poisoned once and the labels remain constant throughout the model's training.

\begin{myitemize}
    \item Random Label Flipping~(RLF~\cite{ECAI:XiaoXE12}): Each poisoned sample is assigned a random class label, i.e., for~$\textit{num}\_\textit{classes}$ classes in the dataset, define the new label as $\textit{new}\_\textit{label} = \texttt{randint}(0, \textit{num}\_\textit{classes} - 1)$.
    \item Targeted Label Flipping~(TLF~\cite{ESORICS:TolpeginTGL20}): Simply flips all labels from a source class to a target class. In \FLname, we always set the source class as~0 and the target class as~1.
    \item Static Label Flipping~(SLF~\cite{USENIX:FangCJG20, SP:SHKR22}): Uses a fixed permutation to determine the new label for each poisoned sample, given by the equation $\textit{new}\_\textit{label} = \textit{num}\_\textit{classes} - \textit{old}\_\textit{label} - 1$.
    \item Dynamic Label Flipping~(DLF~\cite{SP:SHKR22}): Utilizing a surrogate model to flip labels for each sample, our implementation aggregates data from all malicious clients to train a model with the same architecture as the one used in \FLname{} training. Once trained, this model is utilized for inference and the labels are assigned to the least probable output determined by the surrogate model. The term~\enquote{dynamic} is used as the labels rely on the trained model, and by altering the training configuration, the poisoned labels will be modified accordingly. The exact training settings of the surrogate model are given in~Tab.~\ref{tab:dlf_setting}.
    
    \begin{table}[htb!]
     \centering
     \captionsetup{font=small}
     \caption{Training parameters for surrogate model in DLF.}
     \label{tab:dlf_setting}
     \small
     \centering
     \begin{tabular}{rrrrrr}
        \toprule
        Parameter & Epochs & \makecell[r]{Batch\\Size} & \makecell[r]{Learning\\Rate} & Momentum & \makecell[r]{Weight\\Decay} \\
        \midrule
        Values & \numprint{50} & \numprint{128} & \numprint{0.05} & \numprint{0.9} & \numprint{0.0005}\\
        \bottomrule
     \end{tabular}
\end{table}
\end{myitemize}

\subsection{New Robust Aggregation Scheme}
\label{sec:tm}
The most common~FL aggregation scheme, \enquote{\mbox{FedAvg}}, simply computes a~(weighted) average of all inputs~(cf.~\secref{sec:related_work_fl}). In contrast, \emph{robust} aggregation schemes detect and exclude outliers, thus a suitable mitigation against data poisoning. From the schemes surveyed in~\cite{SP:SHKR22,DLS:GMSSWY23,ARXIV:XFG24}, we identify~\enquote{FLTrust}~\cite{NDSS:CaoF0G21} and~\enquote{Trimmed Mean}~(TM)~\cite{ICML:YinCRB18} as the ones with most efficient MPC implementations.

FLTrust~\cite{NDSS:CaoF0G21} measures the cosine similarity between the inputs of participants and the most recent global model trained with a clean data set, then excludes the least similar inputs.
Trimmed Mean~(TM)~\cite{ICML:YinCRB18}, for each coordinate, computes the mean across the provided gradient updates and excludes the values that deviate the most in either direction of the mean.
Performing~TM aggregation obliviously in~MPC requires implementing costly sorting to determine the ranking in each coordinate.

We observe that, intuitively, data poisoning in contrast to model poisoning does not result in specific coordinates producing extreme outliers.
Hence, we propose a heuristic~\enquote{Trimmed Mean Variant} that computes the mean and ranking only for a small~\emph{randomly sampled subset} of coordinates.
Then, during aggregation, it excludes those inputs that occurred the most as outliers in the sample.
We detail our algorithm in~Alg.~\ref{alg:algorithm-tr}, where the underlying MPC functionalities~$\MPCTrimmedMeanList$ and~$\MPCTopKHitter$~(cf.~\tabref{tab:mpc-functionalities}) are as follows:

\begin{algorithm}[htb!]
   \begin{small}
   \caption{Our Trimmed Mean~(TM) Variant in \FLnameTitle{}}
   \label{alg:algorithm-tr}
    \begin{algorithmic}[1]
       \INPUT $\FLmodelSet{}=\{\FLmodel{i}{}\}_{i\in [\numclusters]}$, $\TRParam$, $\TRSampleSize$, $\FLmodelSize$ \COMMENT{$\FLmodelSize = |\FLmodel{i}{}|$, $\TRParam$ -- trim threshold, $\TRSampleSize$ -- sample size}
       \OUTPUT $\MPCSharing{\WWFLmodel{AGG}}{}$ \COMMENT{aggregated model after removing outliers} 
       \STATE {\bfseries initialize:} $\OutlierSet \leftarrow \emptyset$ \COMMENT{set of outliers}
       \STATE $\MPCIndexList \leftarrow$ sample random $\TRSampleSize$ indices from $[1,\FLmodelSize]$.
       \STATE $\myMPCSet{U} \leftarrow \MPCTrimmedMeanList(\FLmodelSet{\MPCIndexList}, \TRParam)$ \COMMENT{$\FLmodelSet{\MPCIndexList}$--Truncated~$\FLmodelSet{}$ with only indices in~$\MPCIndexList$, $\MPCTrimmedMeanList$ performs Trimmed Mean algorithm and returns $2\TRParam$ outlier values~(top and bottom $\TRParam$) for each index in $\MPCIndexList$, $|\myMPCSet{U}| = 2\TRParam\TRSampleSize$}
       \STATE $\myMPCSet{V} \leftarrow \MPCTopKHitter(\myMPCSet{U}, 2\TRParam)$ \COMMENT{returns list of~$2\TRParam$ indices that occur most frequently in~$\myMPCSet{U}$}
       \FORALL{$i \in \myMPCSet{V}$}
            \STATE{$\OutlierSet \leftarrow \OutlierSet \bigcup \{\FLmodel{i}{}\}$}
       \ENDFOR
       \STATE $\MPCSharing{\WWFLmodel{AGG}}{} \leftarrow \MPCAgg(\FLmodelSet{}\setminus\OutlierSet)$
    \end{algorithmic}
    \end{small}
\end{algorithm}

$\MPCTrimmedMeanList$ takes as input a set of vectors, say~$\myMPCSet{W}$, consisting of~$\TRSampleSize$-sized vectors of the form~$\FLmodel{i}{j}$ for~$i \in [\TRSampleSize], j \in [\sizeFL{\myMPCSet{W}}]$.
Moreover, the values in the vector come from a fixed source~(the Layer~II MPC clusters in our case) and are thus represented as a tuple of the form~$\FLmodel{i}{j} = (u_i,v_i)_j$.
Here, $u_i$ denotes the source~ID~(MPC cluster in \FLname) and~$v_i$ represents the corresponding value.
W.l.o.g., consider the first index position of these vectors~($i=1$). $\MPCTrimmedMeanList$ sorts the list~$\{(u,v)_j\}_{j \in [\sizeFL{\myMPCSet{W}}]}$ using the value~$v$ as the key and selects the IDs~($u$) associated with the top and bottom~$\TRParam$ values. 
Intuitively, the operation results in selecting the~MPC clusters whose local updates fall in either the top-$\TRParam$ or bottom-$\TRParam$ position among all the updates at that index.
This procedure is performed in parallel for all~$\TRSampleSize$ indices and results in a set~$\myMPCSet{U}$ of~$2\TRParam\TRSampleSize$ IDs~(with duplicates). 
The~$\MPCTopKHitter$ functionality, parameterized by~$\Gamma$, takes this set~$\myMPCSet{U}$ as input and returns a set of~$\Gamma$ values that occur most frequently in~$\myMPCSet{U}$.

\subsection{CrypTen Implementation Details for Trimmed Mean}
\label{sec:crypten_implementation}
CrypTen lacks a built-in oblivious sorting functionality, so we implement privacy-preserving sorting for so-called~CrypTensors.
Sorting is necessary to compute trimmed mean and our optimized trimmed mean variant. 
We minimize the number of comparisons by implementing a bitonic sorting network that generates the necessary comparisons between elements to allow for a non-stable sorting algorithm.
For trimmed mean, it is not necessary to preserve the relative order of same valued keys as each of them would have been seen as suspicious anyway.
The comparisons are stored in plaintext, as they do not reveal any information about the data.
The comparison steps are only computed once and then executed in parallel for each coordinate.
For~100 elements, we perform~1077 comparisons and for~10 elements only~31.
As the result of each comparison is hidden, we perform the swap operation for each pairs as described in~Lst.~\ref{listing:crypten-sorting}.

\lstset{style=python, escapechar=|}
\begin{figure}[htb!]
\begin{lstlisting}[language=python, caption=CrypTen: Parallel Sorting., label=listing:crypten-sorting]
compare_indices = list(comparision_generator(tensors.size(1)))
for (i, j) in compare_indices:
  b = tensors[:, i].gt(tensors[:, j])
  h = b * tensors[:, i] + (1 - b) * tensors[:, j]
  l = b * tensors[:, j] + (1 - b) * tensors[:, i]
  tensors[:, i] = l
  tensors[:, j] = h
\end{lstlisting}
\vspace{-6mm}
\end{figure}

When computing the proposed trimmed mean variant, we are only interested in the indices of the outliers and therefore only perform the swap operations on a list of indices.
This is done to minimize the compute operations and thereby reduce time and communication.
After identifying which gradients were most often detected as outliers and then computing the set of benign indices, we finally compute the sum over those while preserving privacy.
The procedure is shown in~Lst.~\ref{listing:sumofbenign}.

\lstset{style=python, escapechar=|}
\begin{figure}[htb!]
\begin{lstlisting}[language=python, caption=CrypTen: Sum of Benign Updates., label=listing:sumofbenign]
def bincount(self, indices, num_bins):
    indices_list = crypten.cryptensor(torch.arange(0, num_bins))
    counts = crypten.stack([indices for _ in range(num_bins)], dim=1).eq(indices_list).sum(dim=0)
    return counts[None, :]

def exclude_topk_indices(self, indices, k):
    sorted_indices = self.sort(indices).reshape(-1)
    n = sorted_indices.shape[0]
    return sorted_indices[:n - k]

benign_indices = self.exclude_topk_indices(
    self.bincount(outlier_indices, num_bins=n),
    k=self.exclude_topk)

all_indices = crypten.cryptensor(torch.arange(0, n).repeat(benign_indices.size(0), 1)).t()
eq_mask = all_indices.eq(benign_indices)
agg_sum = gradients.matmul(eq_mask).sum(dim=1)

return agg_sum / (n - self.exclude_topk)
\end{lstlisting}
    \vspace{-6mm}
\end{figure}

Tab.~\ref{tab:aggregation_run_config} shows the parameters used by the aggregation server.
The trimmed mean threshold~$\TRParam$ has been chosen such that it can exclude all malicious updates on either side in the~FL setting.
The worst-case scenario we consider is a poison rate of~0.2, which is equivalent to approx.~20 malicious clients selected per round.

\begin{table}[htb!]
\centering
\captionsetup{font=small}
\caption{Robust aggregation parameters.}
\label{tab:aggregation_run_config}
\begin{small}
\begin{tabular}{llr}
\toprule
Parameter & FL & \FLname \\
\midrule
    Trimmed Mean $\TRParam$ & 20 & 2\\
    Trimmed Mean Variant excluded gradients & 40 & 4\\
    FLTrust root data set size & \multicolumn{2}{c}{200}\\
\bottomrule
\end{tabular}
\end{small}
\end{table}

FLTrust~\cite{NDSS:CaoF0G21} uses a root data set to calculate the server model.
To be most effective, the server model must be similar to the benign models and therefore the root dataset must be representative for the whole dataset. 
Therefore, we sample~(like for all clients)~200 data points for that dataset.

\subsection{Evaluation}
\label{sec:attack-evaluation}
We now empirically evaluate the impact of data-poisoning attacks on~\FLname considering different~(robust) aggregation schemes. For this, we implement all four attacks---RLF~\cite{ECAI:XiaoXE12}, TLF~\cite{ESORICS:TolpeginTGL20}, SLF~\cite{USENIX:FangCJG20, SP:SHKR22}, DLF~\cite{SP:SHKR22}---in our framework and add~CrypTen-based implementations of three robust aggregation schemes to our prototype implementation.
Using the setup described in~\secref{sec:evaluation}, we answer the following questions next:
\begin{myitem}
    \setcounter{enumi}{3}
    \item [Q5:] What is the impact of data-poisoning attacks on the accuracy of~FL and~\FLname using~FedAvg and robust aggregation schemes?
    \item [Q6:] What is the run-time and communication overhead for different robust aggregation schemes in~MPC?
    \item[Q7:] How does our~TM variant compare to regular~TM~w.r.t.\ accuracy and~MPC performance?
\end{myitem}

\begin{figure*}[htb!]
    \centering
    \includegraphics[width=\textwidth]{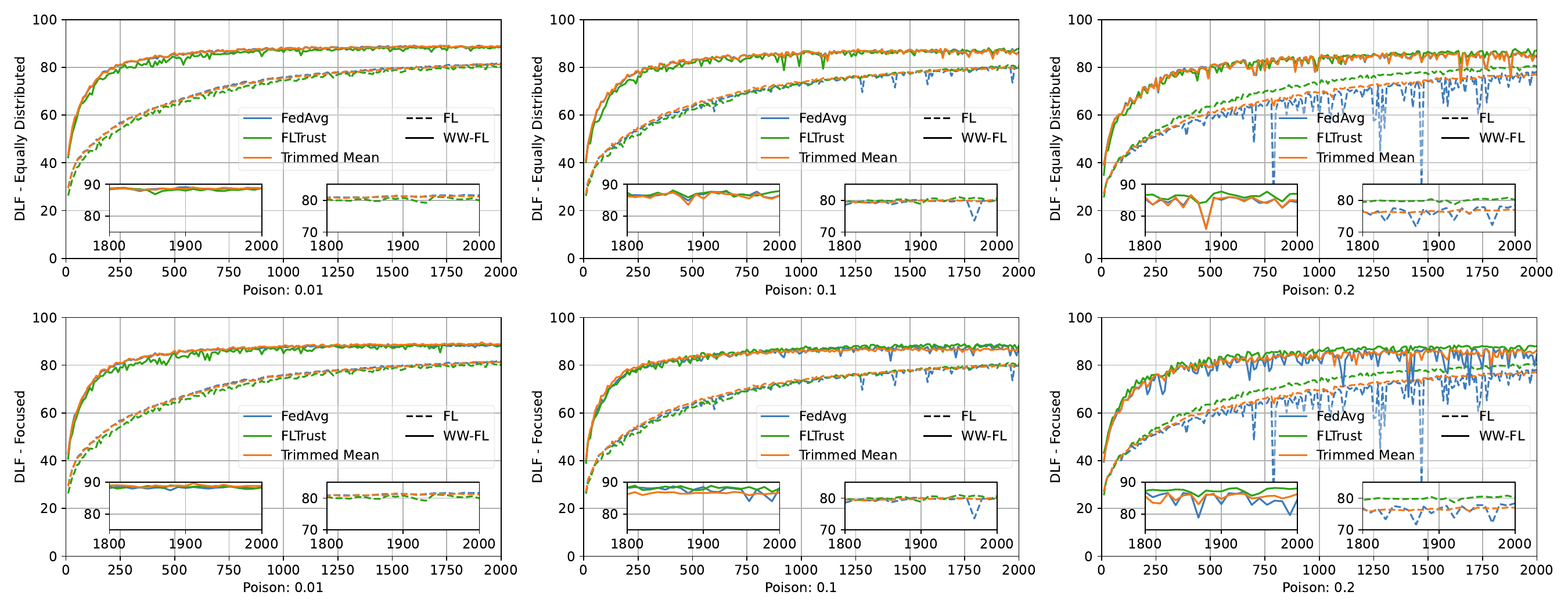}
    \vspace{-8mm}
    \captionsetup{font=small}
    \caption{Validation accuracy for~FL and~\FLnameTitle{} when training~ResNet9 on~CIFAR10 with~FedAvg, FLTrust, and~trimmed mean as aggregation schemes under~DLF attack for three different poison rates~(top: equally distributed, bottom: focused setting).}
    \label{fig:main_CIFAR10-FL-HyFL-FedAvg-FLTrust-TrimmedMean-2000-DLF-attack-focused-equally-distributed}
\end{figure*}

\myparagraph{Q5 -- Attack Impact}
We consider three poison rates~(0.01, 0.1, 0.2) and distributions of attackers:
in the~\emph{equally-distributed} setting, we assume that on expectation each cluster has the same number of malicious clients; 
in the~\emph{focused} setting, we assume that malicious clients are concentrated on as few clusters as possible while there is still an honest majority in each cluster~(a standard assumption in~FL); 
finally, in the unrealistic~\emph{cluster-focused} setting, we see what happens if we lift the honest-majority assumption and concentrate all malicious clients in as few clusters as possible.
In~Fig.~\ref{fig:main_CIFAR10-FL-HyFL-FedAvg-FLTrust-TrimmedMean-2000-DLF-attack-focused-equally-distributed}, we study the effectiveness of the most powerful~DLF~\cite{SP:SHKR22} attack on both regular~FL and~\FLname when training~ResNet9 on~CIFAR10 in the equally distributed and focused setting. Additional experimental outcomes, including evaluation over the cluster-focused setting, are given in~\secref{sec:app-additional-benchmarks}.

For the fairly aggressive poison rate of~0.2, we see in both visualized attacker distributions a significant negative impact of the~DLF attack on~FL when using~FedAvg with drops below~30\% accuracy.
However, these can be successfully mitigated with robust aggregation schemes.
While there is also negative impact on~\FLname, especially in the focused setting, the accuracy even with~FedAvg never drops below that of~FL.
Even though robust aggregation schemes help to slightly smoothen the curve, we conclude that~\emph{applying defenses in~\FLname against data-poisoning attacks is optional, but not strictly necessary}.

\myparagraph{Q6 -- Robust Aggregation in MPC}
We evaluate the run-time and communication overhead of our~FLTrust and~TM implementation in~CrypTen in~Tab.~\ref{tab:crypten_aggregation}. The run-time overhead for both robust aggregation schemes compared to~Fed\-Avg is four to five orders of magnitude higher~(ranging from~5 to~30 minutes).
Also, FLTrust requires~5$\times$ more run-time and communication than~TM.
Given that both produce fairly similar results when applied to~\FLnameTitle{}, the overhead for~FLTrust seems not justified.

\begin{table}[htb!]
\centering
\captionsetup{font=small}
\caption{Communication (Comm. in MB) and run-time (Time in seconds) for various aggregation schemes, including our Trimmed Mean (TM) variant with sample sizes~10, 100, and~1000 in~CrypTen.}
\label{tab:crypten_aggregation}
\begin{small}
\begin{tabular}{lrrrrrr}
\toprule
Param & FedAvg & FLTrust & TM & TM-10 & TM-100 & TM-1000\\
\midrule
    Comm. & 0 & \numprint{5329.19} & \numprint{1021.59} & \numprint{9.69} & \numprint{11.90} & \numprint{33.94}\\
    Time  & \numprint{0.023} & \numprint{1713.44} & \numprint{326.57} & \numprint{558.05} & \numprint{558.26} & \numprint{561.50} \\
\bottomrule
\end{tabular}
\end{small}
\vskip -0.1in
\end{table}

\begin{figure}[htb!]
    \centering
    \includegraphics[width=0.49\textwidth]{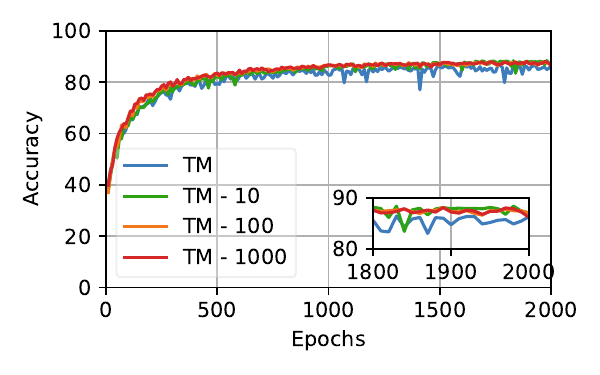}
    \vspace{-3mm}
    \captionsetup{font=small}
    \caption{Accuracy of trimmed mean~(TM) and our variant~(with sample sizes~10, 100, and~1000) against focused~DLF attacks on~\FLname at~0.2 poison rate for~ResNet9/CIFAR10.}
    \label{fig:main_HyFL-TrimmedMean-variant-2000-DLF-focused}
    \vspace{-2mm}
\end{figure}

\myparagraph{Q7 -- Trimmed Mean Variant}
In~Fig.~\ref{fig:main_HyFL-TrimmedMean-variant-2000-DLF-focused}, we compare the effectiveness of our~TM variant to the original~TM~\cite{ICML:YinCRB18} for three sample sizes~(10, 100, and~1000).
It turns out that our heuristic approach barely reduces the effectiveness, even with aggressive parameters.
In fact, in the focused setting, the~TM variant outperforms the original.
This is because our variant completely excludes updates of~(poisoned) outliers, whereas in regular trimmed mean, those poisoned updates might still be considered for some coordinates.
Results for all other settings are presented in~\secref{sec:app-additional-benchmarks}.

In~Tab.~\ref{tab:crypten_aggregation}, we provide run-times and communication results for our optimizations.
Compared to the original with~1.02GB of communication, we can see an improvement by two orders of magnitude for the variant with~100 random samples.
However, we see a higher and fairly stable run-time across all three examined variants.
This is because the algorithm for determining the overall ranking of outliers increases the number of~MPC communication rounds. 
In the studied inter-continental~WAN setting, this has severe impact but does not correspond to actual compute time.
Overall, if~\FLname is combined with a robust aggregation scheme, our~TM variant offers an excellent trade-off between accuracy and~MPC overhead.

\paragraph{Additional Benchmarks}
We provide additional results in~\secref{sec:app-additional-benchmarks}, including additional accuracy comparisons of \FLname{} with standard~FL in all three modes of corruption, and the validation accuracy results for our Trimmed Mean variant.
\section{Conclusion}
\label{sec:conclusion}
In this work, we presented~\FLname, a novel unified abstraction and framework for large-scale~(hierarchical) federated learning that provides global model privacy, faster convergence, smaller attack surface, and better resilience against poisoning attacks than regular~FL.
We envision that~\FLname will lead to a paradigm shift in~FL system design and deployment.
Future efforts can extend our foundational work with more efficient instantiations, stronger security, and comprehensive evaluations with larger datasets and setups.
\section*{Acknowledgments}
\label{sec:acknowledgments}
This project received funding from the European Research Council~(ERC) under the European Union's research and innovation programs Horizon~2020 (PSOTI/850990) and Horizon Europe (PRIVTOOLS/101124778). It was co-funded by the Deutsche Forschungsgemeinschaft~(DFG) within SFB~1119 CROSSING/236615297.

\bibliographystyle{alpha}
\bibliography{bibliography}

\appendix

\section{Additional Benchmarks}
\label{sec:app-additional-benchmarks}
In this section, we present additional benchmarks and results to further evaluate the performance of~\FLname. These supplementary findings provide valuable insights into the effectiveness of~\FLname in comparison to regular federated learning~(FL) across various scenarios and attack settings.

\subsection{Additional Results for Q4}
\label{sec:app:CIFAR10-attacks}

\begin{figure}[htb!]
    \centering
    \begin{sideways}
    \includegraphics[width=1.3\linewidth]{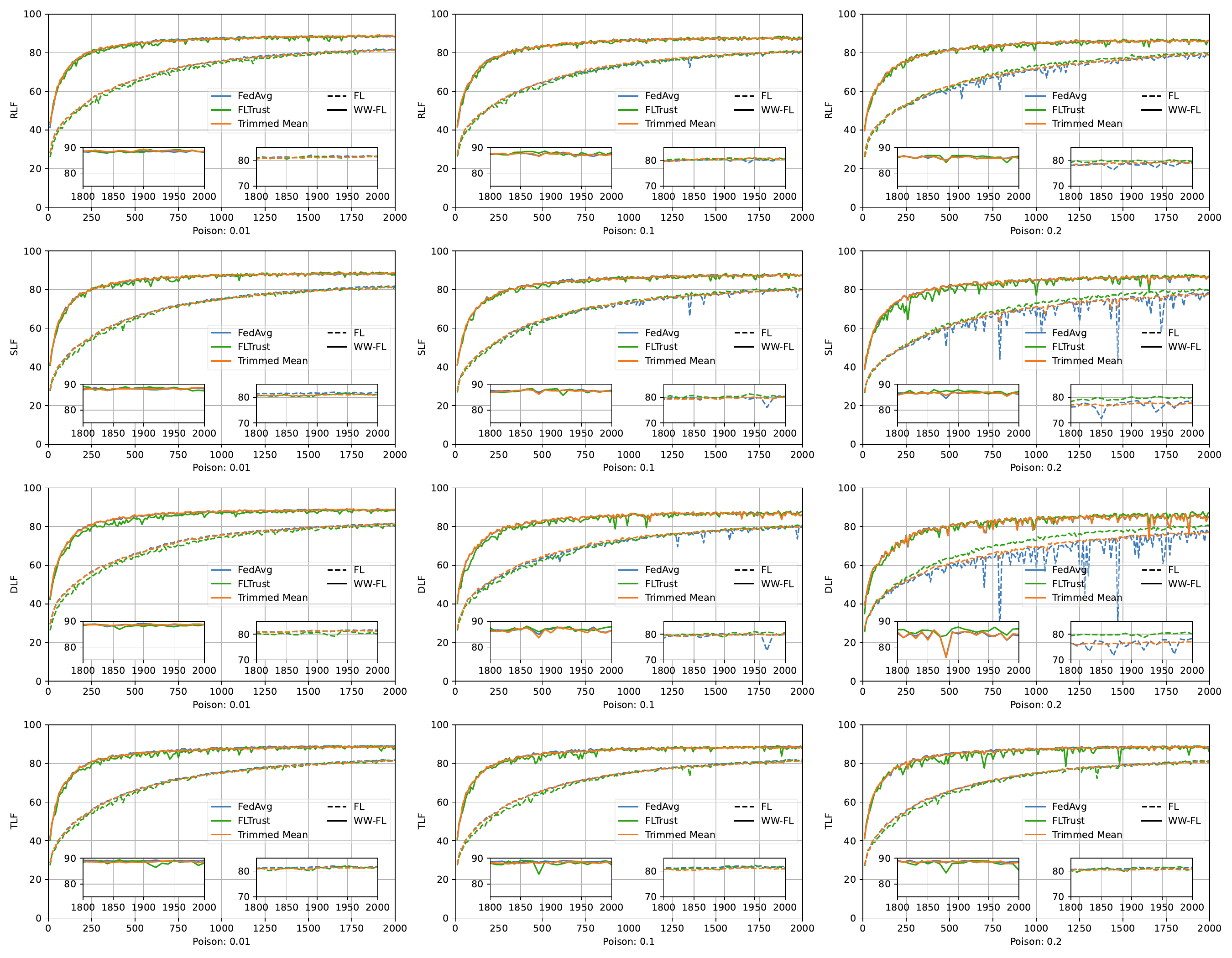}
    \caption{Validation accuracy for~ResNet9/CIFAR10 training with~FedAvg, FLTrust, and~trimmed mean for~2000 iterations under~RLF, SLF, DLF, and~TLF attacks in the equally-distributed setting.}
    \label{fig:CIFAR10-FL-HyFL-FedAvg-FLTrust-TrimmedMean-2000-all-attack-equally-distributed}
    \end{sideways}
    \vspace{-6mm}
\end{figure}

\begin{figure}[htb!]
    \centering
    \begin{sideways}
    \includegraphics[width=1.3\linewidth]{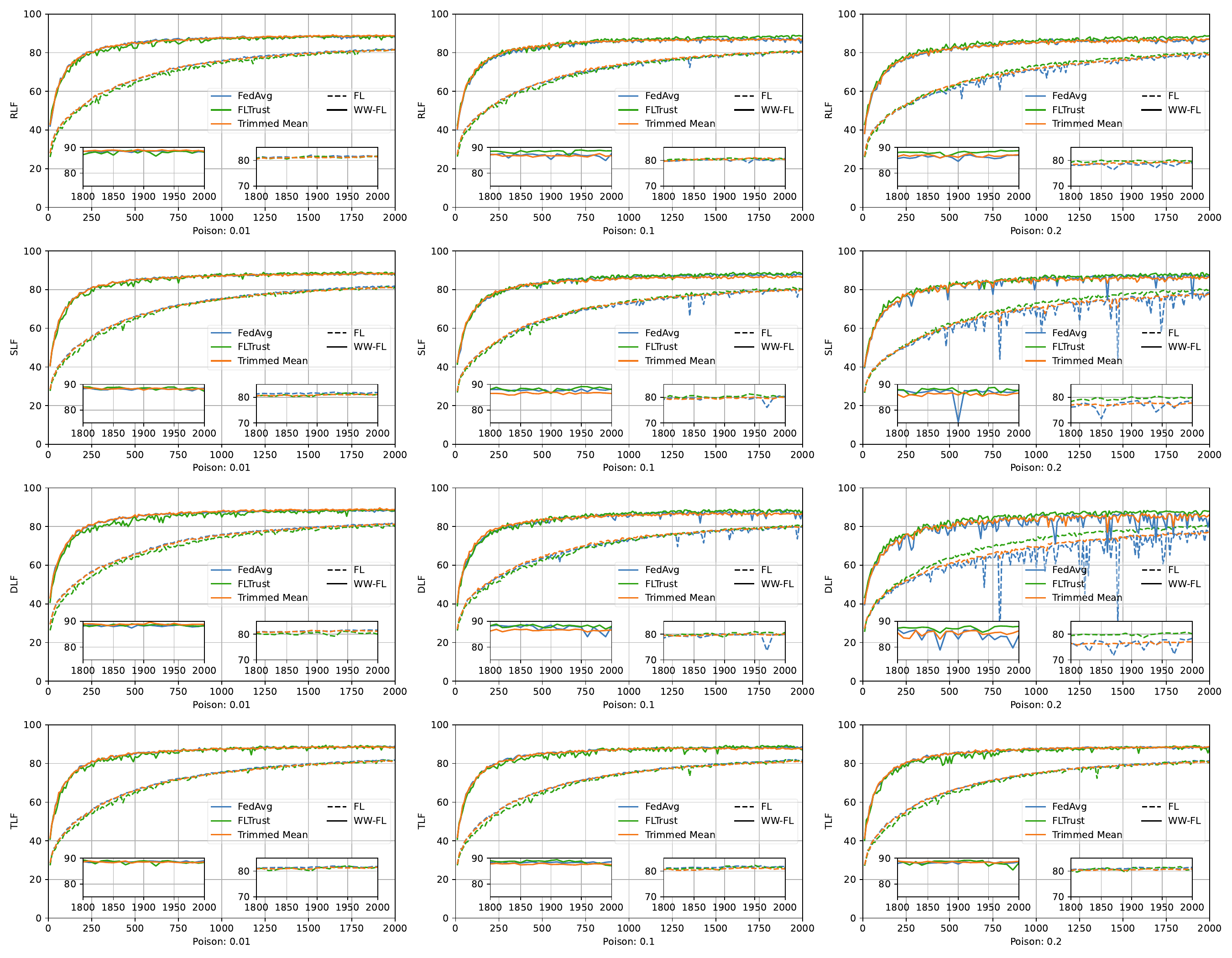}
    \vspace{-3mm}
    \caption{Validation accuracy for~ResNet9/CIFAR10 training with~FedAvg, FLTrust, and~trimmed mean for~2000 iterations under~RLF, SLF, DLF, and~TLF attacks in the focused setting.}
    \label{fig:CIFAR10-FL-HyFL-FedAvg-FLTrust-TrimmedMean-2000-all-attack-focused}
    \end{sideways}
\end{figure}

\begin{figure}[htb!]
    \centering
    \begin{sideways}
    \includegraphics[width=1.3\linewidth]{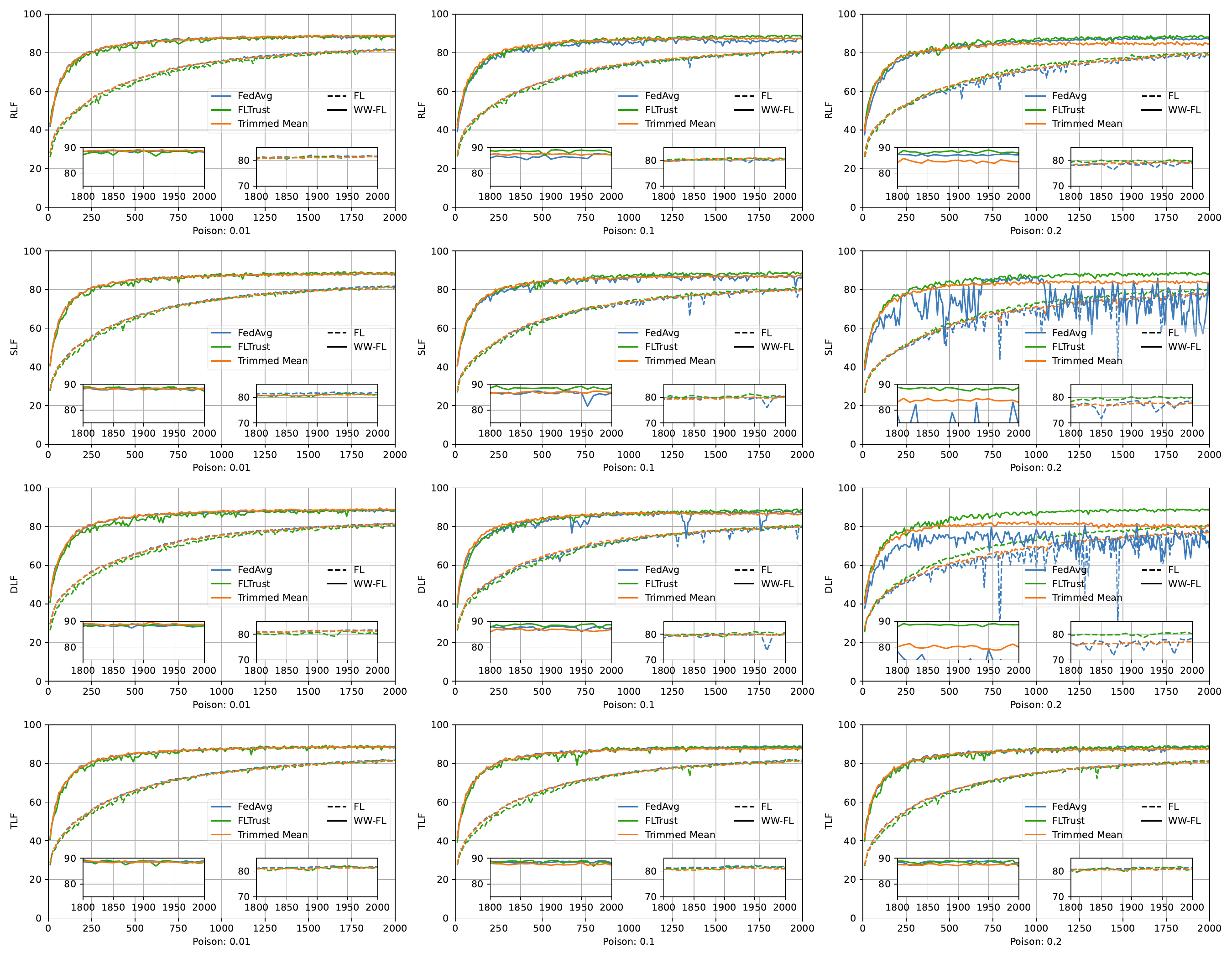}
    \vspace{-3mm}
    \caption{Validation accuracy for~ResNet9/CIFAR10 training with~FedAvg, FLTrust, and~trimmed mean for~2000 iterations under~RLF, SLF, DLF, and~TLF attacks in the cluster-focused setting.}
    \label{fig:CIFAR10-FL-HyFL-FedAvg-FLTrust-TrimmedMean-2000-all-attack-cluster-focused}
    \end{sideways}
\end{figure}

\paragraph{ResNet9/CIFAR10}
Here, we present the full evaluation results for~ResNet9 trained on~CIFAR10 under attack.
We evaluate~FedAvg, FLTrust, and trimmed mean for~2000 rounds in regular~FL and~\FLname under four different data-poisoning attacks with three different poison rates and three different distributions of malicious clients. Fig.~\ref{fig:CIFAR10-FL-HyFL-FedAvg-FLTrust-TrimmedMean-2000-all-attack-equally-distributed} shows the results for equally-distributed malicious clients, Fig.~\ref{fig:CIFAR10-FL-HyFL-FedAvg-FLTrust-TrimmedMean-2000-all-attack-focused} the focused attack, and~Fig.~\ref{fig:CIFAR10-FL-HyFL-FedAvg-FLTrust-TrimmedMean-2000-all-attack-cluster-focused} cluster-focused results.

\FLname outperforms regular~FL in nearly all settings and has more than~5\% higher validation accuracy after~2000 rounds.
In both regular~FL and~\FLname, all three aggregation schemes show robustness against data poisoning in the realistic settings with~0.01 and~0.1 poison rate and equally-distributed as well as focused attacks.
This is not the case for a poison rate of~0.2.
Especially~FedAvg as a non-robust aggregation scheme struggles to converge.
Even robust aggregations cannot fully counteract the attacks.
The same is true for the cluster-focused attack distribution:
Here, all aggregations have spikes in their validation accuracy; only~FLTrust manages to train a model comparable to the other settings.
This is because~FLTrust is designed to be robust against model poisoning, and the cluster-focused setting is the closest to a model-poisoning attack.
We stress that the cluster-focused scenario is less practical in reality. Consider worldwide training that is organized into country-level clusters: it is extremely unlikely that an adversary can corrupt even the majority of a country's population. Corruption in a production-level cross-device setting is below a rate of~0.001~\cite{SP:SHKR22}, making our study of extreme cases with a~0.2 corruption rate a theoretical experiment.


\begin{figure}
    \centering
    \begin{sideways}
    \includegraphics[width=1.3\linewidth]{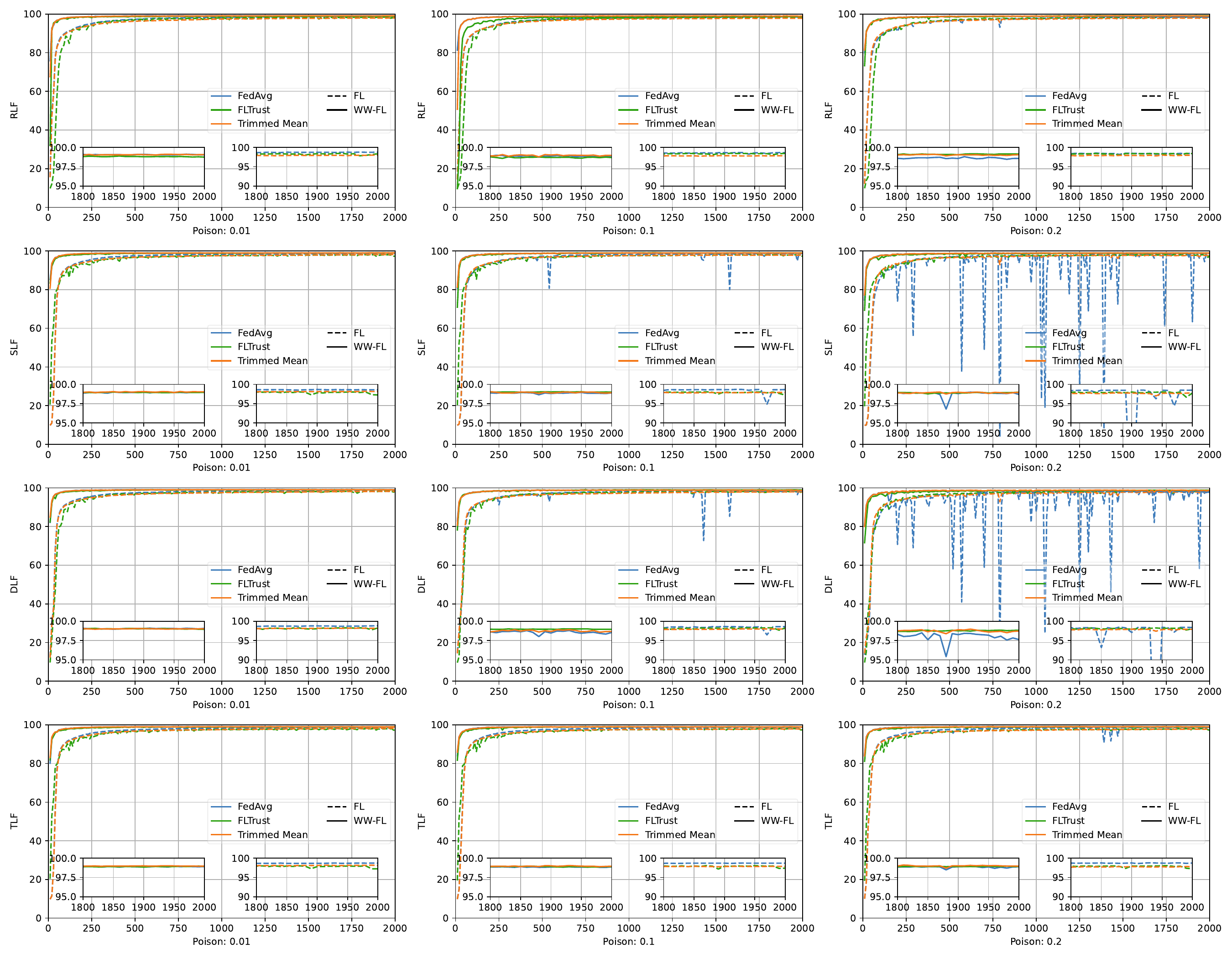}
    \caption{Validation accuracy for~LeNet/MNIST training with~FedAvg, FLTrust, and~trimmed mean for~2000 iterations under~RLF, SLF, DLF, and~TLF attacks in the equally-distributed setting.}
    \label{fig:MNIST-FL-HyFL-FedAvg-FLTrust-TrimmedMean-2000-all-attack-equally-distributed}
    \end{sideways}
\end{figure}

\begin{figure}
    \centering
    \begin{sideways}
    \includegraphics[width=1.3\linewidth]{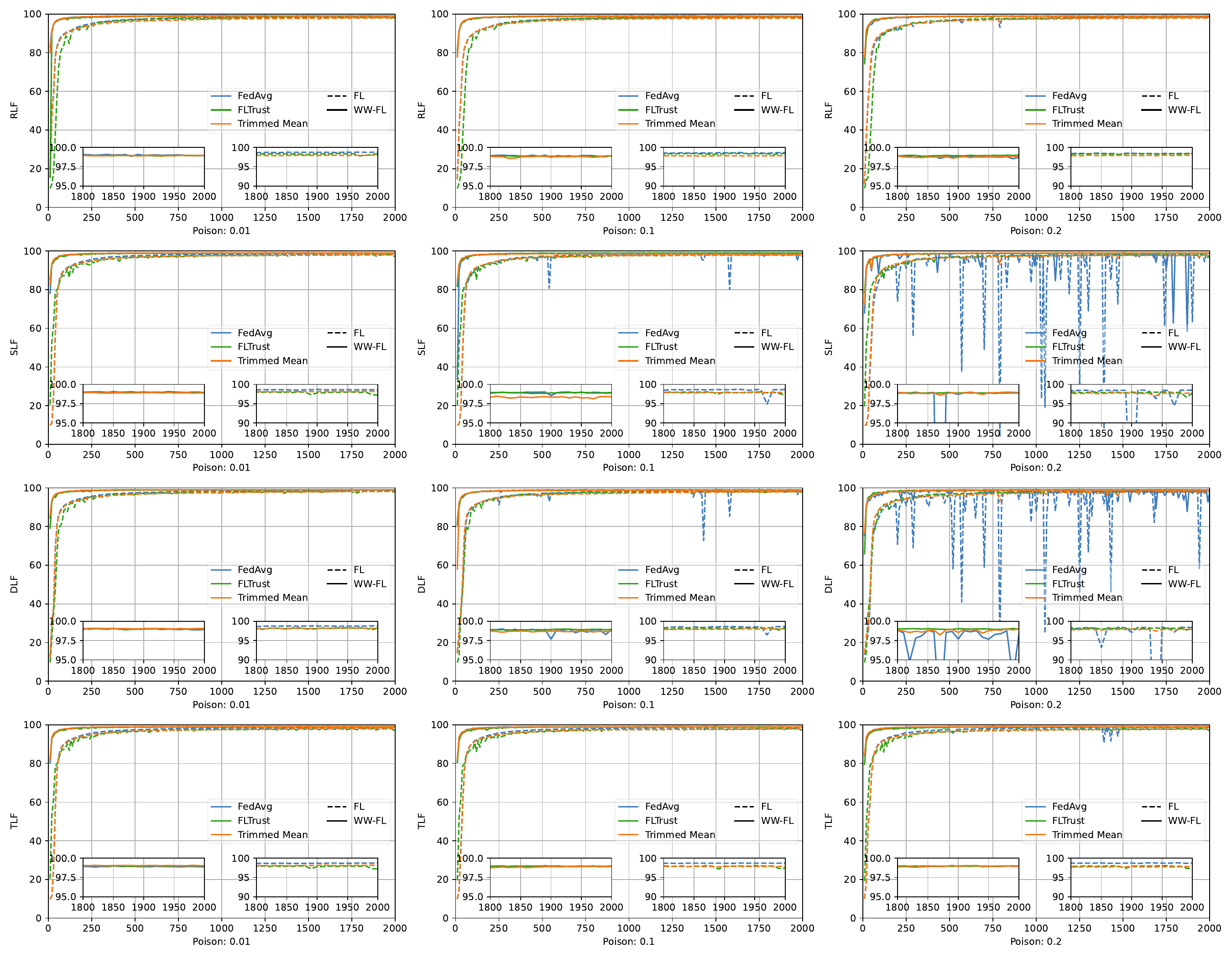}
    \caption{Validation accuracy for~LeNet/MNIST training with~FedAvg, FLTrust, and~trimmed mean for~2000 iterations under~RLF, SLF, DLF, and~TLF attacks in the focused setting.}
    \label{fig:MNIST-FL-HyFL-FedAvg-FLTrust-TrimmedMean-2000-all-attack-focused}
    \end{sideways}
\end{figure}

\begin{figure}
    \centering
    \begin{sideways}
    \includegraphics[width=1.3\linewidth]{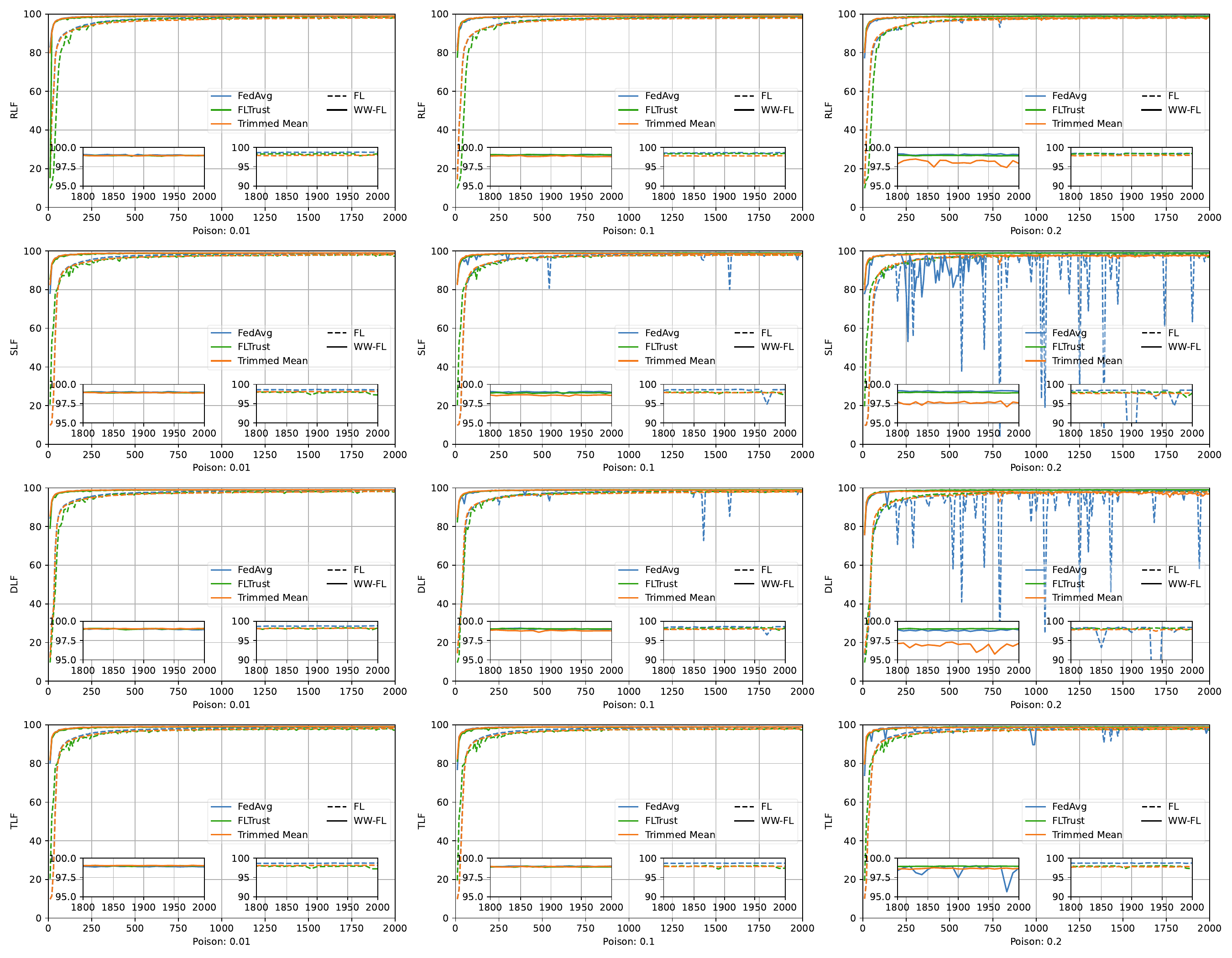}
    \caption{Validation accuracy for~LeNet/MNIST training with~FedAvg, FLTrust, and~trimmed mean for~2000 iterations under~RLF, SLF, DLF, and~TLF attacks in the cluster-focused setting.}
    \label{fig:MNIST-FL-HyFL-FedAvg-FLTrust-TrimmedMean-2000-all-attack-cluster-focused}
    \end{sideways}
\end{figure}

\begin{figure}[htb!]
    \centering
    \begin{sideways}
    \includegraphics[width=1.3\linewidth]{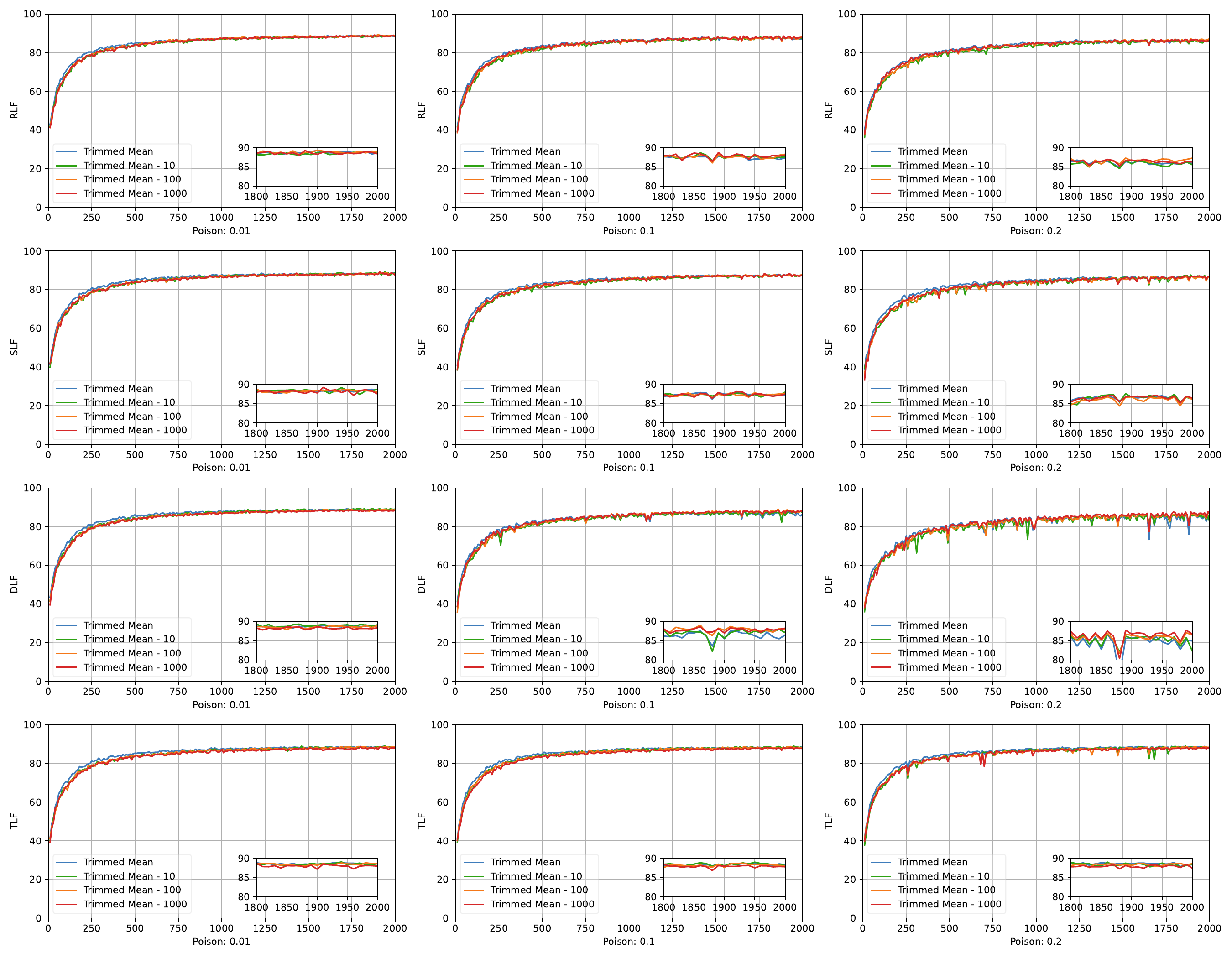}
    \caption{Validation accuracy for~ResNet9/CIFAR10 training with~trimmed mean and our~trimmed mean variant for~2000 iterations under~RLF, SLF, DLF, and~TLF attacks in the equally-distributed setting.}
    \label{fig:CIFAR10-HyFL-TrimmedMean-Variant-2000-all-attack-equally-distributed}
    \end{sideways}
\end{figure}

\begin{figure}[htb!]
    \centering
    \begin{sideways}
    \includegraphics[width=1.3\linewidth]{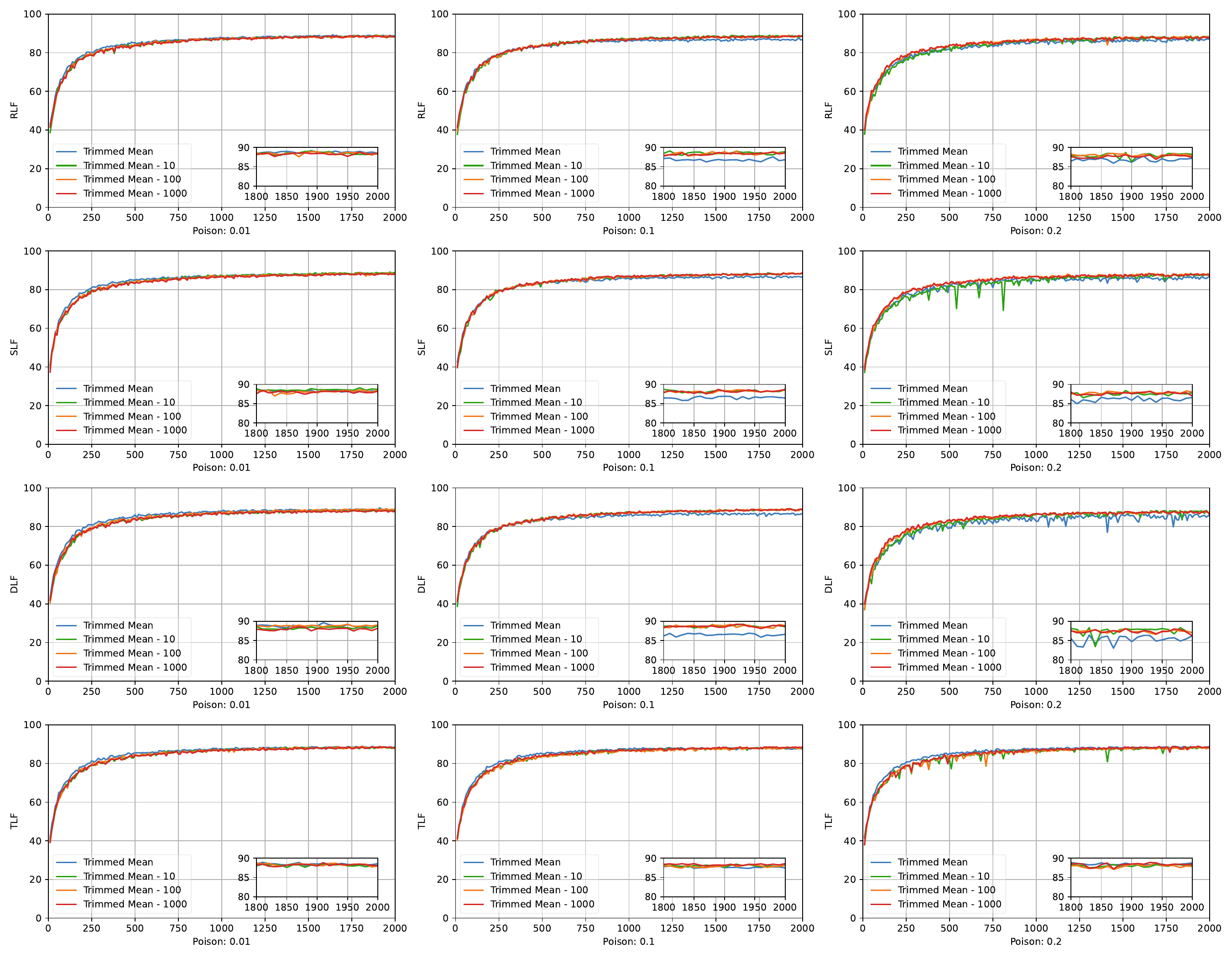}
    \caption{Validation accuracy for~ResNet9/CIFAR10 training with~trimmed mean and our~trimmed mean variant for~2000 iterations under~RLF, SLF, DLF, and~TLF attacks in the focused setting.}
    \label{fig:CIFAR10-HyFL-TrimmedMean-Variant-2000-all-attack-focused}
    \end{sideways}
\end{figure}

\begin{figure}[htb!]
    \centering
    \begin{sideways}
    \includegraphics[width=1.3\linewidth]{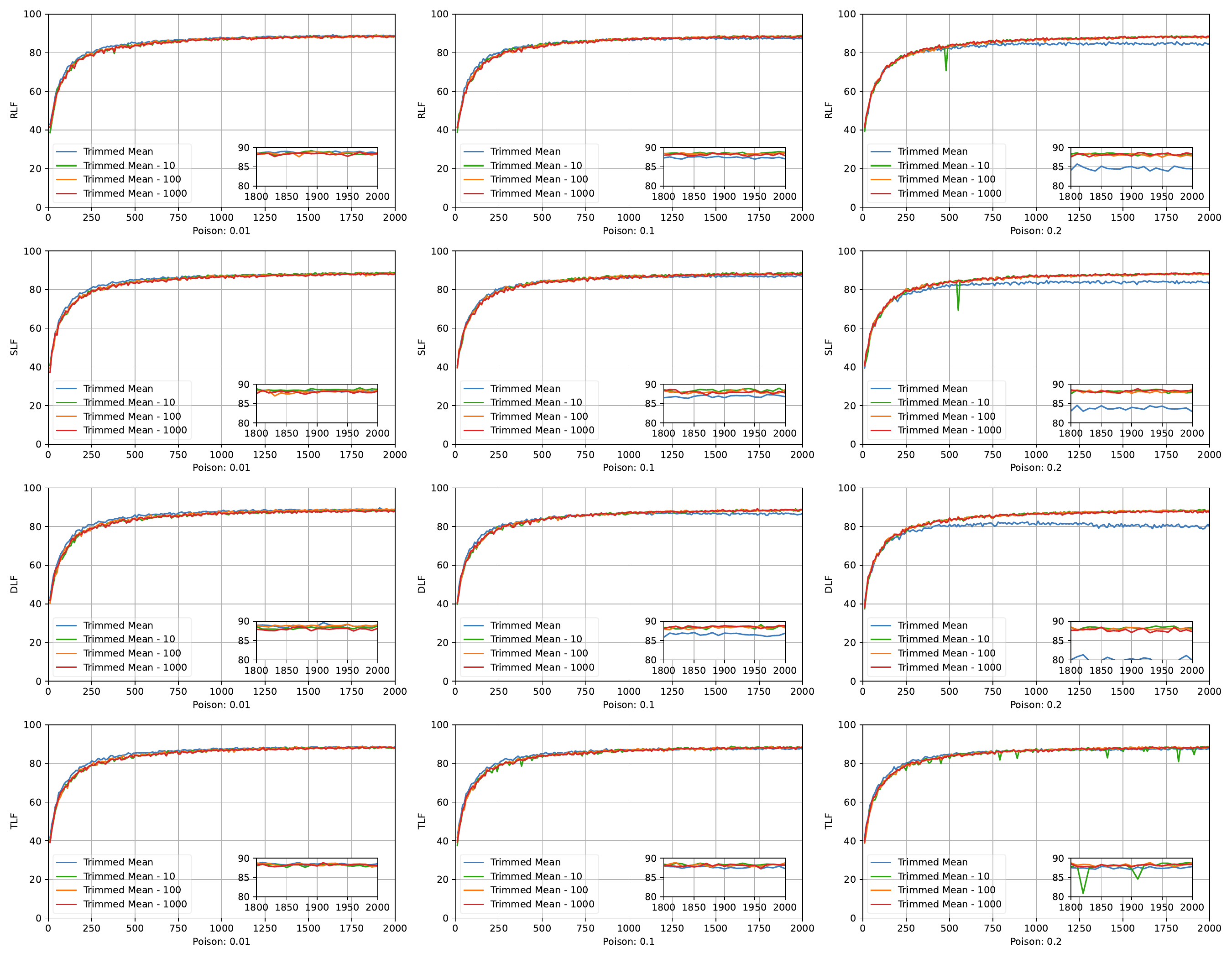}
    \caption{Validation accuracy for~ResNet9/CIFAR10 training with~trimmed mean and our~trimmed mean variant for~2000 iterations under~RLF, SLF, DLF, and~TLF attacks in the cluster-focused setting.}
    \label{fig:CIFAR10-HyFL-TrimmedMean-Variant-2000-all-attack-cluster-focused}
    \end{sideways}
\end{figure}


\paragraph{LeNet/MNIST}
We show additional evaluation results for~LeNet trained on~MNIST for~2000 iterations in~Fig.~\ref{fig:MNIST-FL-HyFL-FedAvg-FLTrust-TrimmedMean-2000-all-attack-equally-distributed}~(equally-distributed), Fig.~\ref{fig:MNIST-FL-HyFL-FedAvg-FLTrust-TrimmedMean-2000-all-attack-focused}~(focused), and~Fig.~\ref{fig:MNIST-FL-HyFL-FedAvg-FLTrust-TrimmedMean-2000-all-attack-cluster-focused}~(cluster-focused).
As for ResNet9, \FLname trains the model faster than regular~FL, but both converge to a similar accuracy in most cases after~2000 rounds.
Here, the advantage of~\FLname over regular~FL mostly lies in the faster convergence: after only~100 rounds of training, \FLname nearly reaches~98\% accuracy, regardless of the attack or attack setting.
Also, there is little difference between the three aggregation schemes for the realistic settings.
For the~0.2 poison rate, FedAvg starts to struggle a lot to train the model.
This can mostly be seen for~FedAvg in regular~FL and sometimes in~\FLname.

\subsection{Additional Results for Q6}
\label{sec:app:Q6-res}
~
\paragraph{ResNet9/CIFAR10}
We present the extended evaluation of our trimmed mean variant for~CIFAR10 when sampling~10, 100, and~1000 coordinates under all attacks and attack settings.
Again, the plots show the evaluation results for~2000 iterations and are divided into three distributions for malicious clients:
Fig.~\ref{fig:CIFAR10-HyFL-TrimmedMean-Variant-2000-all-attack-equally-distributed} shows equally-distributed, Fig.~\ref{fig:CIFAR10-HyFL-TrimmedMean-Variant-2000-all-attack-focused} shows focused, and~Fig.~\ref{fig:CIFAR10-HyFL-TrimmedMean-Variant-2000-all-attack-cluster-focused} shows cluster-focused.

In most cases, all four aggregations perform equally and the final validation accuracy is nearly the same. The only difference can be seen with the cluster-focused attack distribution and~0.2 poison rate. Here, trimmed mean achieves a substantially lower accuracy than the trimmed mean variant. We assume that is because cluster-focused acts more like a model poisoning attack and trimmed mean variant deals better with this by discarding whole gradients rather than operating coordinate-wise.

\paragraph{LeNet/MNIST}
We also compared trimmed mean against our trimmed mean variant for~LeNet being trained on~MNIST.
However, almost all settings, the aggregations perform very similar.
Hence, we omit the full plots.

\end{document}